\documentclass[bst/sn-mathphys]{sn-jnl}% Math and Physical Sciences Reference Style
\usepackage{etoolbox}
\makeatletter
\patchcmd{\ps@headings}
{\hbox to \hsize{\hfill Springer Nature 2021 \LaTeX\ template\hfill}}
{\hbox to \hsize{}}
{}
{}
\patchcmd{\ps@headings}
{\hbox to \hsize{\hfill Springer Nature 2021 \LaTeX\ template\hfill}}
{\hbox to \hsize{}}
{}
{}
\patchcmd{\ps@titlepage}
{\hbox to \hsize{\hfill Springer Nature 2021 \LaTeX\ template\hfill}}
{\hbox to \hsize{}}
{}
{}
\makeatother

\usepackage{url}
\usepackage[T1]{fontenc}
\usepackage{graphicx}
\usepackage[round]{natbib}
\usepackage{pifont}
\usepackage{float}

\newcommand\sdiam{\ding{169}}

\jyear{2022}%

\theoremstyle{thmstyleone}%
\theoremstyle{thmstyletwo}%

\theoremstyle{thmstylethree}%

\begin{document}

\title[An Attention-Based Model for Contextual Informativeness]{An Attention-Based Model for Predicting Contextual Informativeness and Curriculum Learning Applications}


\author*[1]{\fnm{Sungjin} \sur{Nam}}\email{sungjin.nam@act.org}

\author[2]{\fnm{David} \sur{Jurgens}}\email{jurgens@umich.edu}
\author[3]{\fnm{Gwen} \sur{Frishkoff}}\email{gfrishkoff@gmail.com}
\author[2]{\fnm{Kevyn} \sur{Collins-Thompson}}\email{kevynct@umich.edu}

\affil*[1]{
% \orgdiv{Research}, 
\orgname{ACT, Inc}
% \orgaddress{\street{500 ACT Dr.}, \city{Iowa City}, \postcode{52243}, \state{Iowa}, \country{USA}}
}

\affil[2]{
% \orgdiv{School of Information}, 
\orgname{University of Michigan}
% \orgaddress{\street{105 S State St.}, \city{Ann Arbor}, \postcode{48109}, \state{Michigan}, \country{USA}}
}

\affil[3]{
% \orgdiv{Psychology Department}, 
\orgname{University of Oregon}
% \orgaddress{\street{1030 East 13th Ave}, \city{Eugene}, \postcode{97403}, \state{Oregon}, \country{USA}}
}


\abstract{
Both humans and machines learn the meaning of unknown words through contextual information in a sentence, but not all contexts are equally helpful for learning. We introduce an effective method for capturing the level of \textit{contextual informativeness} with respect to a given target word.
% , based on attention models in deep learning. 
Our study makes three main contributions. 
First, we develop models for estimating contextual informativeness, focusing on the instructional aspect of sentences. 
% models for estimating contextual informativeness using pre-trained embeddings and an attention layer. 
Our attention-based approach using pre-trained embeddings demonstrates state-of-the-art performance on our single-context dataset and an existing multi-sentence context dataset. 
Second, we show how our model identifies key contextual elements in a sentence that are likely to contribute most to a reader's understanding of the target word. 
Third, we examine how our contextual informativeness model, originally developed for vocabulary learning applications for students, can be used for developing better training curricula for word embedding models in batch learning and few-shot machine learning settings. 
We believe our results open new possibilities for applications that support language learning for both human and machine learners.
}




\maketitle

\section{Introduction}
We learn new vocabulary with significant help from context. Humans acquire the meanings of unknown words partially and incrementally by repeated exposure to clues in the surrounding text or conversation~\citep{frishkoff2008measuring}. 
As part of literacy training, contextual word learning methods can help students by teaching them different techniques for inferring the meaning of unknown words by recognizing and exploiting semantic cues such as synonyms and cause-effect relationships~\citep{heilman2010personalization}. 

However, not all contexts are equally informative for learning a word's meaning. As Figure \ref{fig:intro-example} shows, there can be wide variation in the amount and type of information about a \textit{target} word to be learned, via semantic constraints implied by the context. 
Humans are very good at few-shot learning of new vocabulary from such examples, but the instructional \emph{quality} of initial encounters with a new word is critical. 

Identifying the degree and nature of contextual informativeness in authentic learning materials is an important problem to solve for designing effective curricula for contextual word learning~\citep{webb2008effects}. 
As we elaborate in Section~\ref{sec:related_work}, predicting and characterizing contextually informative passages for learning is quite different from other context-based prediction tasks such as n-gram prediction or cloze completion. 
For example, some contexts are better than others for learning because they provide more effective support for inferring the \textit{meaning} of the target word. Generic natural language processing models may not capture this target-specific contextual informativeness, which is a critical factor in determining instructive quality in contextual word learning applications.
In this paper, we investigate an approach for automatically identifying and characterizing contexts with varying levels of support for learning a given target word. Further, we show this approach also has broad potential applications for both human and machine learning.

\begin{figure}[t]
\centering
\begin{tabular}{ | p{0.94 \linewidth} | } 
\hline
% \small
1) My friends, family, and I all really like \textit{tesg\"uino}. \\
% \small
2) There is a bottle of \textit{tesg\"uino} on the table. \\
% \small
3) Brewers will ferment corn kernels to make \textit{tesg\"uino.} \\
\hline
\end{tabular}
\caption{
Three sentences that have the same length but provide very different contextual information about the meaning of the target word, \textit{tesg\"uino.} We introduce a new dataset and computational models to quantify the degree and nature of this target-specific \textit{contextual informativeness} for learning.}
\label{fig:intro-example}
\end{figure}

We introduce examples of predicting the degree and understanding the nature of the \textit{contextual informativeness} of a passage with respect to the meaning of a target word to be learned. We also present its applications in developing training curricula for machine learning models. 
% Our study makes three main contributions.
First, we show that recent advances in deep semantic representations are highly effective for this task.
We demonstrate that our model based on BERT~\citep{devlin2019bert}, combined with a masked attention layer, generalizes effectively across very different datasets, giving state-of-the-art performance not only for our single-sentence context dataset but also on a previous multi-sentence context dataset of~\cite{kapelner2018predicting}. 
Second, beyond predicting a score, we also provide quantitative and qualitative evaluations of how our model captures the contributions of a particular passage to correctly infer a target word's meaning, demonstrating that the masked attention activation provides fine-grained, interpretable characterizations of contextual informativeness across various semantic relations. 
% Further, using the dataset of~\cite{santus2015evalution}, we show that informativeness learned through this mechanism is robust across various semantic relations. 
Third, we investigate how the contextual informativeness prediction model for designing a vocabulary learning curriculum for human learners can be also useful for word embedding models in multiple machine learning scenarios. 
We believe our results are applicable not only to developing educational curricula for vocabulary instruction, but also to NLP tasks like few-shot machine learning of new words or concepts from text~\footnote{Data and code are shared at \url{https://github.com/sungjinnam/contextual_informativeness}}.

%%%%%%%%%%%%%%%%%%%%%%%%%%%%%%%%%%%%%%%%%%%%%%%%%%%%%%%%%%%%%%%%%%%%%%%%%%%%%%%%%%%%%%%%%%%%%%%%%%%%%%%%%%%%%%%%%%%

\section{Related Work}
\label{sec:related_work}
Our study is connected to important prior work in literacy research, including modeling contextual informativeness, natural language processing, and curriculum machine learning. As background, our study focuses on measuring \textit{contextual informativeness} with respect to a specific target word. This has some connection to the predictability of the word, or the ``likelihood of a word occurring in a given context'' in psycholinguistics~\citep{warren2012introducing}, but with important differences that we describe further below. Generic definitions of informativeness, not associated with a specific target word, have been defined in multiple ways, e.g., as the density of relevant information that exists in a dialog~\citep{gorman2003evaluation}, or the number of different semantic senses included in a sentence~\citep{lahiri2015squinky}. Entropy-based measures like \textit{KL-divergence} have been used to represent readers' surprise from reading new texts compared to their prior knowledge~\citep{peyrard2018formal}. Compared to these generic definitions, the contextual aspect relative to a target word has a critical distinction: different words in the same sentence may have very different degrees of semantic constraint imposed by the rest of the sentence. Further, computational lexical semantics has long studied how to characterize word meaning in context~\citep{mikolov2013distributed} and how contextual information can be used to select word meaning~\citep{szarvas2013supervised,melamud2016context2vec,mccann2017learned,peters2018deep,devlin2019bert}. 
However, these models typically assume informative contexts are given to solve the task, and do not predict or characterize the varying degrees of informativeness with respect to a target word. 

\subsection{Contextual Informativeness in Literacy Research}
\label{sec:related__context_literacy}
% - contextual informativeness: quality of context w.r.t. the target word
% - contextual word learning examples 
% - role of different informative sentences
Educational research has examined multiple ways in which context influences reader's ability to learn and retain word meaning. For example, \cite{beck1983vocabulary} characterized the informativeness of contexts for learning new words, distinguishing between pedagogical (specifically chosen to teach meaning) vs. natural (unintentionally informative). They reviewed two basal reading series and found that sentences fell into four contextual informativeness categories: misdirective (actually leading to erroneous learning: about 3\% of observations); non-directive (ambiguous sentences with little information value about the target word: 27\%); general (sentences that help place the target word in a general category: 49\%); and directive (sentences that happen to point the student to the target word's specific, correct meaning: 13\%). In our study, the terms non-directive, general, and directive map to low, medium, and high contextual informativeness respectively. We omit the misdirective case for now given its relative rarity.

In general, repetitive exposure to such contextual cues from text or conversation can provide much information about the meaning of unknown words~\citep{frishkoff2008measuring}. Contextual word learning is an instructional method that teaches students how to infer the meaning of unknown words by recognizing and utilizing semantic cues, such as synonyms and cause-effect relationships~\citep{heilman2010personalization}. More recent research has shown that both high- \emph{and} low-informative contexts play important roles in optimizing long-term retention of new vocabulary, as they invoke different but complementary learning mechanisms. Low-informative contexts force more retrieval from memory, while high-informative contexts elicit a variety of inference processes that aid deeper word comprehension~\citep{van2018contextual}. Exposing a reader to the right carefully-chosen curriculum of different contexts can lead to significantly better long-term retention of new words. 
For example, \cite{frishkoff2016dynamic} showed that using a scaffolded series of informative contexts (e.g., initially highly informative, then progressively less informative) resulted in the best long-term retention of new words (+15\%), compared to several other curriculum designs. 
Since different levels of informativeness can elicit different learning behaviors, it is important to distinguish different levels of informative learning materials in contextual word learning. 
% We now discuss how prior work used machine learning to find contextually informative material.
% Similar results were obtained later by~\citep{van2018contextual}. 

% Humans learn better when the examples presented for learning are organized in a meaningful way. 
%For example,~\cite{frishkoff2016accuracy} showed that using an ordered series of informative contexts (e.g., starting from highly informative, followed by less informative examples) provided better long-term retention of new words than several other curriculum designs. Since different levels of informativeness can elicit different learning behaviors (e.g., low-informative contexts facilitate retrieval from memory, while high-informative contexts can promote various inference processes and comprehension \citep{van2018contextual}) it is important to distinguish different levels of informative learning materials in contextual word learning.

\subsection{Models for Contextual Informativeness}
\label{sec:related__models_informativeness}
% - difference from language model's output
Only one previous study, to our knowledge, has done substantial work on using machine learning to characterize the contextual informativeness of curriculum materials for vocabulary learning. The study by~\cite{kapelner2018predicting} explored using predictive models based on random forests that combined over 600 different pre-specified text features. They applied their technique to their own dataset of multi-sentence text passages, which we include in our analysis.

Our approach differs from theirs in several significant ways. 
First, instead of requiring large numbers of hand-coded features, our approach learns effective feature representations \emph{automatically} using attention-based deep learning. We show that this allows our model to generalize more robustly to new datasets, without requiring extensive feature engineering. We also find that combining our approach with selected features from theirs into a hybrid model attains the best overall performance. 
Second, we explore the \emph{interpretability} of the resulting models, to characterize \emph{how} a particular context gives information about a given target word. 
The complex and large number of engineered features from ~\cite{kapelner2018predicting} can become difficult to obtain clear explanations of prediction results on individual contexts.
Third, their specific focus was primarily on achieving high-precision prediction for the most highly informative contexts from authentic online texts. 
In contrast, we focus on predicting \emph{and} characterizing contexts across a \emph{range} of low- and high-informative levels. In Section~\ref{sec:contextual_informativess_model}, 
% we describe our deep learning-based model that addresses these issues and attains improved performance.
we describe the deep learning-based model used for our curriculum construction that addresses these issues and attains improved performance, while achieving even better performance on a benchmark dataset. 

In earlier work, the REAP project~\citep{CollinsThompson2004} used NLP methods to identify appropriate contexts for vocabulary learning, but focused on filtering entire web pages by tagging sentences with specific criteria, not individual prediction of informative contexts. 
Similarly, another study used a feature-engineering approach with supervised learning to develop a classifier to label entire documents as learning objects for concepts (e.g., computer science)~\citep{hassan2008learning}, but did not focus on quantifying or characterizing the informativeness of context passages for specific target words.
In Sections~\ref{sec:exp1} and \ref{sec:exp2}, we show that our model can successfully estimate the degree of contextual informativeness of individual contexts, and provides interpretable outcomes for the prediction results.

\subsection{Contextual Models in Natural Language Processing}
The predictability of a word given its surrounding context is often represented as a probability calculated from a large corpus~\citep{jurafsky2001probabilistic}. Language models, in particular, can provide useful information on which words may come after the given context.
However, language modeling alone may not adequately capture semantics for contextual informativeness. Additional longer-range dependencies, or more sophisticated semantic relations and world knowledge may be needed~\citep{shaoul2014cloze}. Also, standard retrieval techniques, such as beam search, may not suitable for getting semantically diverse responses from language models~\citep{vijayakumar2018diverse}.

Models like ELMo~\citep{peters2018deep} use LSTM layers~\citep{hochreiter1997long} to capture semantic information from a sequence of words. 
Transformer-based models like BERT~\citep{devlin2019bert} can be also used to represent contexts that consist of word sequences. 
Unlike LSTM-based models, the latter can be more effective in understanding long-range dependencies or unusual relationships between words~\citep{khandelwal2018sharp}. 
Both models can be considered as a bi-directional neural language model since they capture the conditional probability of the target word based on the sentential context. 

In this paper, we use two contextual pre-trained embeddings, ELMo~\citep{peters2018deep} and BERT~\citep{devlin2019bert} as baselines and compare the performance in our model. 
Although these contextual representation models can learn rich information about the word's meaning based on context words, it is often hard to achieve generalizable interpretation of what does these relationships actually mean~\citep{clark2019does}. Similarly, predicting entropy of predicted cloze responses may not represent the instructional quality of a given context. 
Our work focuses on the \textit{instructional} aspect, which investigates whether and how context words facilitate making correct inferences about the \textit{meaning} of the target word~\citep{beck1983vocabulary}. 
% Our suggested model was inspired in part by~\cite{liu2018content}. They used an attention-based model to classify customer sentiment towards particular product aspects, by capturing the relationship between context words and a target word.
% We develop a model that can predict the degree of instructional information on contextual word learning examples, and identify the instructional relationship between context words and the target word. 

% In contrast, we are interested in the continuum of potential informative contexts as well as characterizing the information provided by a given context.
% Finally, we also contribute a significant new labelled dataset for contextual informativeness prediction, based on single sentences and high quality labels estimated with an ensemble of labelling methods.

% In our study, we examined how an attention mechanism used on top of a pre-trained embedding model can predict the amount of contextual information with respect to the unknown target word. 

\subsection{Related Tasks in Natural Language Processing}
\label{Sec:related__similar_tasks}
Various NLP tasks examines the meaning of a word in a particular context. 
First, several lexical tasks seek to produce acceptable responses based on a context. 
% First, several lexical tasks focus on predicting acceptable words for a given informative context. 
Lexical substitution tasks have the model choose the correct word that can replace an existing word in a sentence, which can be shown once per sentence~\citep{mccarthy2007semeval} or multiple times~\citep{kremer2014substitutes}. 
Lexical completion tasks, like the Microsoft Sentence Completion Challenge~\citep{zweig2011microsoft}, have the model fill in the blank without providing an example target word. 
However, our task aims to predict the degree of informativeness of a context, assuming the amount of contextual information can vary depending on the selected target word. 

Second, previous studies proposed tasks for predicting the semantic properties of predetermined concepts~\citep{wang2017distributional} or named entities~\citep{pavlick2017identifying}.
However, they assume that particular semantic senses for target concepts or entities are presented in the training set. 
Our annotated task focuses on having a model predict the degree of semantic constraint in a single- or multi-sentence context without using predefined lists of concepts for evaluation. 

Third, nonce word tasks include various learning scenarios for unseen words. 
Previous studies investigated how contextual information can be used to infer the meaning of synthetically generated target words~\citep{lazaridou2017multimodal,herbelot2017high}. 
However, they also relied on the assumption that the provided context contains enough information to make an inference, by manually selecting the training sentences for synthetic words. Each synthetic \textit{Chimera} word was generated by combining the context (e.g., sentence) of two or more words that already exist in the training examples. 
In contrast, our contextual informative task involves diverse examples where some contexts can be less or more helpful. Our model also attempts to characterize the nature of the explicit cues that exist for learning the target word.

\subsection{Curriculum Learning in Natural Language Processing}
\label{sec:related__curriculum_learning}
% - loss based
% - ours: contextual informativeness (human word learning) based
Previous curriculum learning studies on machine learning models have focused on achieving more efficient training~\citep{bengio2009curriculum}.
For example, multiple NLP studies investigated the role of training curriculum on performance. 
\cite{sachan2016easy} showed that easier (e.g., smaller loss) but more diverse training examples (e.g., with different locations in semantic feature space) 
can provide better performance for NLP models when solving question-answering tasks. 
The properties of the learning target, such as topical domain or part-of-speech, may also affect the stability of word embedding models~\citep{wendlandt2018factors}.  
Particular context words can be more important than others for predicting the target word (e.g., entropy of the next word's probability), and weighting each context word differently can improve training efficiency~\citep{kabbach2019towards}. We also note that \cite{swayamdipta2020dataset} recently suggested that machine learning algorithms may also benefit more generally from datasets with different informativeness levels, such as high-informative examples for improving model optimization and less-informative examples for model generalization.

% Unlike these studies, our study focuses on the idea that if applying vocabulary learning strategies for helping human students can also benefit machine learning models.
% By using a model that can predict contextual informativeness, we score sentences from the corpus and developed simple filtering heuristics for producing higher-quality training curricula. We also examine how contextual informativeness scores can be used for more efficient training of word embedding models. 
Instead of exploring generic curriculum learning strategies, our study examines whether vocabulary learning strategies similar to those that benefit human students could also benefit machine training for language-related tasks. As we describe later, one approach we implement is to score sentences from the corpus using our contextual informativeness model, develop simple filtering heuristics for producing higher-quality training curricula, and examine how the contextual informativeness score-based curricula provides more efficient training of word embedding models.

%%%%%%%%%%%%%%%%%%%%%%%%%%%%%%%%%%%%%%%%%%%%%%%%%%%%%%%%%%%%%%%%%%%%%%%%%%%%%%%%%%%%%%%%%%%%%%%%%%%%%%%%%%%%%%%%%%%

\section{Attention-Based Model for Predicting Contextual Informativeness}
% - BERT + attention: why chose attention? -- interpretable contribution of contexts to informativeness w.r.t. the target word
% - BERT + attention + lexical features (for multi-sentence dataset)
% - TODO: other model structure? e.g., attention flow: use it like attention weights
\label{sec:contextual_informativess_model}

In the following experiments, we introduce and evaluate an attention-based model that predicts contextual informativeness scores (Sections~\ref{sec:exp1} and \ref{sec:exp2}) and also explore its application in developing curricula for machine learning models (Sections~\ref{sec:exp3} and \ref{sec:exp4}).

\begin{figure}[t]
    \centering
    \includegraphics[width=0.5\linewidth]{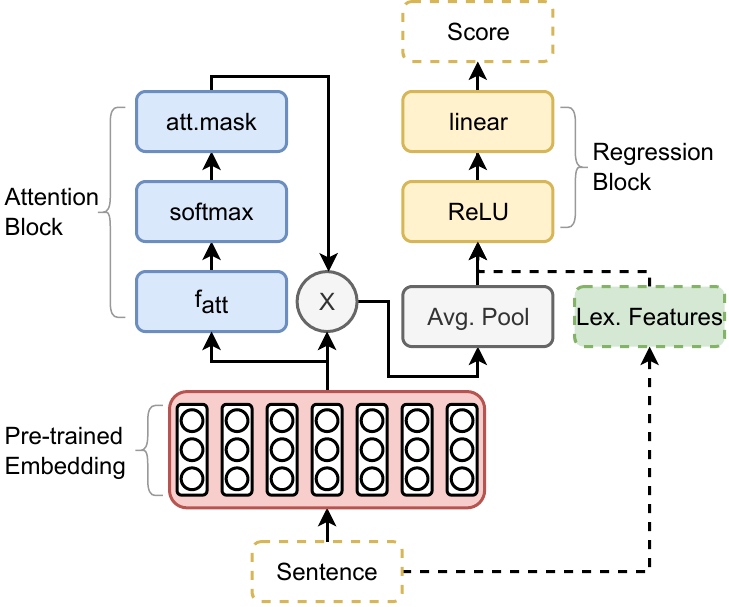}
    % \hfill
    \caption{
    The structure of our proposed model. 
    The model consists of a pre-trained embedding (red) with masked attention block (blue) to create attention weighted context vectors, and regression block (yellow) to predict the numeric contextual informativeness score.
    For the multi-sentence context dataset, we also tested lexical features from~\cite{kapelner2018predicting} (green) as a complementary input.
    % We compared the prediction performance of the model without (left) and with attention block (right). 
    }
    \label{fig:mod_structure}
\end{figure}

Our model predicts contextual informativeness using deep learning methods to represent the context and identify which aspects contribute to identifying a word's meaning. 
We used pre-trained components (red block in Figure \ref{fig:mod_structure}) to retrieve vector representations of contexts and the target word. 
Specifically, we compared the use of pre-trained versions of ELMo~\citep{peters2018deep} and BERT-Base (12 layers, 768 dimensions)~\citep{devlin2019bert} models from Tensorflow Hub in developing our model. %\footnote{\url{https://tfhub.dev/tensorflow/bert_en_cased_L-12_H-768_A-12/1}}.
%pre-trained models from the public repository for each model.

During training, we selectively updated the pre-trained models' parameters to avoid overfitting: for ELMo-based models, we updated parameters that determine the aggregating weights of LSTM and word embedding layers; for BERT-based models, we updated the parameters for the last layer.
For each model, we treated the target word as unknown (e.g.,  \texttt{<UNK>} for ELMo models or \texttt{[MASK]} for BERT models) token so that the model must use contextual information to infer the meaning of the \textit{unknown} target word. 

The input for the attention layers (blue blocks) is the vector for the target word and context tokens.
Using a multiplicative attention mechanism~\citep{luong-pham-manning:2015:EMNLP}, we calculated the relationship between the masked target word and context words ($f_{att}$). We used $softmax$ to normalize the output of the attention layer. The output of the softmax layer masked non-context tokens as zero, to eliminate the weights for padding and the target word ($att.mask$).
The masked attention output was then multiplied with the contextual vectors from the pre-trained model to generate attention-weighted context vectors. This masked attention structure was inspired in part by a prior attention-based model that classified customer sentiment towards particular product aspects by capturing the relationship between context words and a target word~\citep{liu2018content}.

We also explored complementing our model's representation with lexical features from~\cite{kapelner2018predicting} (green block) by concatenating their features with our attention-weighted context vectors. %(e.g., \texttt{BERT+Att+Lex} in Figure \ref{fig:pred_res_multi}). 
The regression layers (yellow blocks) used an average pooling result of attention-weighted context vectors. 
The layers comprised a ReLU layer and a fully-connected linear layer that estimated the score of contextual informativeness on a continuous scale. We used root mean square error (RMSE) as a loss function.
More details for the model are in Appendix~\ref{app:hyperparams}.

%%%%%%%%%%%%%%%%%%%%%%%%%%%%%%%%%%%%%%%%%%%%%%%%%%%%%%%%%%%%%%%%%%%%%%%%%%%%%%%%%%%%%%%%%%%%%%%%%%%%%%%%%%%%%%%%%%%
\section{Datasets: Contextual Informativeness}
\label{sec:exp1__dataset}
% - Existing dataset and discussion of considerations in how that dataset was construction
% - new annotations on existing dataset, with clear highlights of what is new
% - some evaluation of quality beyond inter-annotator agreement?  Measure IAA directly and with split-rank correlation (both were high).
% - Some example here of the resulting data to build intuition. Reiterate how low informative contexts are still useful for education.
In the following experiments, we tested our models trained with two significantly different datasets: single- and multi-sentence context datasets. 
Both datasets were developed for vocabulary learning applications with instructional purposes. 
% We used two different datasets to train and evaluate our model's performance on predicting contextual informativeness.
Each dataset introduces a challenging prediction task with significant room for improvement, even for the best current NLP models like ELMo or BERT. 
However, these datasets have very different attributes. For example, they have different context lengths (single- vs. multi-sentence contexts), number of included contexts (1,783 vs. 67,833), labeling methods (relative vs. absolute assessment scales), range of target words (Tier 2 vs. various levels), and source of sentences (manually crafted vs. web-scraped).
Using datasets with varied characteristics helped us test the generality of our contextual informativeness model in different situations.

Both datasets included gold-standard scores for contextual informativeness: numeric values based on the perceived learning effectiveness of the context for the given target word. These scores effectively summarized the amount of contextual informativeness of a given context, such as the precision and variety of any cues that are present in the context that help a reader infer the precise, correct meaning of the target word. Specific examples of cues might include synonymy, antonymy, cause-effect, whole-part, frequent co-occurrence, or other relationships that help comprehend the meaning of a new word. However, because of the virtually unlimited nature of these cues, for both datasets, annotators were not given explicit relation types as a basis for judgment. 
For the analysis, as single- and multi-sentence context datasets have different score ranges, we conducted min-max normalization for each score type to make them comparable. 
% We now describe the details of these two datasets.

\subsection{Single-Sentence Contexts}

The first dataset is our pedagogical single-sentence contexts for contextual vocabulary learning. 

\label{sec:exp1__dataset_single}
% - Details on crowdsourcing settings
%   - example sentences
%   - how sentences are developed
%   - instruction given to workers
%   - inter-rater agreements
%   - scores distribution; why it can be correlated with sentence length

\begin{table}[t]
\centering
% \small
\begin{tabular}{  p{0.95 \linewidth} } \hline
\textbf{High Informative}: The barks and howls of dogs created too much $\rule{7mm}{0.15mm}$ for us to sleep. \\
% \small 
\textit{Original Target Word:}   din\\
\textit{Avg. Informativeness Score:} 0.75 \\ \hline % 0.913\\ \hline
\textbf{Low Infomative}: We weren't able to tell if he was a $\rule{7mm}{0.15mm}$ or not. \\
% \small 
\textit{Original Target Word:}   recluse\\
\textit{Avg. Informativeness Score:} -0.54 \\ \hline% 0.239\\ \hline
\end{tabular}
\caption{
For the single-sentence dataset, researchers generated high (top), medium, or low (bottom) informative context sentences. 
We additionally crowdsourced numeric annotations for informativeness score (BWS method), to cross-check the informativeness level of each generated sentence (Section~\ref{sec:dataset_dscovar_bws}). 
}
\label{tab:single_sentence_context_ex}
\end{table}

\subsubsection{Generating a Dataset}
These sentences were originally developed for an intelligent tutoring system for contextual vocabulary learning~\cite{frishkoff2016dynamic}\footnote{Dynamic Support of Contextual Vocabulary Acquisition for Reading (DSCoVAR) (\url{http://dscovar.org/}).}.
The system was created and curated by experts in reading and linguistics for use in classroom studies and enrichment programs that help students to improve their literacy skills. This dataset’s quality and effectiveness in a contextual word learning application have been validated in multiple classroom studies with hundreds of students.
For this study, we added annotations across carefully controlled target words and different levels of contextual informativeness.

This annotated data consists of 1,783 sentences. Each sentence contains exactly one target word drawn from a set of 60 words (20 nouns, 20 verbs, and 20 adjectives).
These target words were \textit{Tier 2} words (critical for literacy but rarely encountered in speech), carefully normed to achieve a balanced set of psychometric properties (abstractness or concreteness, age of acquisition, etc.).

With these target words, researchers (not the authors) with a background in literacy research generated sentences according to high, medium, or low informativeness guidelines.
As we noted in Section \ref{sec:related__context_literacy}, contextual vocabulary learning applications can benefit from having different levels of informative sentences. High informative sentences can be useful for inference processes while low informative sentences can elicit retrieval processes (Table \ref{tab:single_sentence_context_ex}). These sentences are designed to deliver more robust contextual word learning results from various levels of contextual informativeness.

The sentences were normed to control variability in semantic and syntactic properties, such as length and difficulty.
The average length of these sentences was 12.49 words ($\sigma^2 = 2.75$ words), and the average relative location of the target word was 64.37\%  from the beginning ($\sigma^2 = 2.83$\%). 
We provided detailed guides for creating these sentences, including examples of different contextual cues and how can they manipulate the amount of contextual informativeness.
Instructions we provided to the researchers to generate sentences can be found in Appendix~\ref{app:sent_gen}.
% Further details are in Appendix~\ref{app:sent_gen}.

\subsubsection{Annotating Perceived Informativeness} 
\label{sec:dataset_dscovar_bws}
For the single-sentence contexts that were generated, we cross-checked the original researcher-provided labels with additional crowdsourced annotations using the best-worst scaling (BWS) method. 
BWS is often preferable to other strategies like ranking with a Likert scale in cases where annotators can reliably distinguish between items (e.g., which sentence is more informative), while keeping the size of annotations manageable. 
% but may have trouble placing them on a specific spot on a scale.
Previous studies have used BWS annotation to create semantic lexicons~\citep{kiritchenko2017best,rouces2018generating}.

We used cloze sentences for the annotation task and asked non-expert crowdworkers to ``find the most- and least-informative sentences'' with respect to the cloze word's meaning,
% to measure context words' informativeness regardless of worker's different knowledge levels on Tier 2 target words. 
We wanted to measure the informativeness of the context itself on the target’s potential meaning, without using annotators’ prior knowledge of Tier 2 target words. This approach was similar to~\cite{lazaridou2017multimodal}, which used synthetic words as target words to measure context sentences’ informativeness.
Before the annotation process, we provided detailed instructions and examples of high and low informative sentences to crowdworkers.

For each question item, workers selected the best or worst informative cloze sentence from a set of four sentences.
The choice of using four-sentence sets is based on previous studies~\citep{kiritchenko2016capturing,kiritchenko2017best} and our own pilot testing. 
Each tuple included four randomly selected different sentences as options to be selected as the best or least informative sentences; each tuple was then scored by 3 different crowdworkers. 
Each annotator scored sentences in -1, 0, or 1. For example, if the sentence was marked as the most informative or the least informative sentence from the question, each rating was converted to an integer score of $+1$ or $-1$ respectively. The unmarked case was considered to be $0$. 
For each sentence, we used the the average score of 24 ratings, collected from three annotators. 

Annotation results for BWS scores were highly reliable. 
Following best practices for measuring annotation replicability as outlined in~\citep{kiritchenko2016capturing}, we simulated whether similar results would be obtained over repeated trials. 
Annotations were randomly partitioned into two sets and then each used to compute the informativeness scores, comparing the rankings of two groups. We repeated this process 10 times, and found the average of Spearman's rank correlation coefficients was 0.843 ($\sigma^2$=0.018, all coefficients were statistically significant ($p < 0.001$)), indicating high replicability in the scores.  
Our replicability scores are in line with previous studies~\citep{kiritchenko2017best,kiritchenko2016capturing}. 
% Inter-rater agreement rates for the best and worst sentence picks for each tuple were 0.376 and 0.424 with Krippendorff's $\alpha$. Although these scores scores are in moderate level, we view this agreement is still acceptable given the replicability score across random splits is similar from the previous studies~\citep{kiritchenko2016capturing,kiritchenko2017best}. 
More details on annotation process can be found in Appendix~\ref{app:score_cloze}.

\subsection{Multi-Sentence Contexts}
\label{sec:exp1__dataset_multi}

\begin{table}[t]
\centering
% \small
\begin{tabular}{  p{0.95 \linewidth} } \hline
\textbf{High Informative}:  As with ginger, turmeric has \textit{salubrious} properties. It is an antiseptic, applied as a paste to cuts and abrasions, and is taken with food to purify the blood and aid digestion. Hindus consider it sacred and the women will mark a dot of it on each other's foreheads as a mark of respect and friendship.\\
% \small 
\textit{Original Target Word:} salubrious\\
\textit{Avg. Score:} 3.22\\ \hline
\textbf{Low Infomative}: The Clandestine Laboratory Training Facility is located at Camp Upshur on the Marine Corps Base in Quantico, Virginia. With the increase in the number of \textit{clandestine} laboratory seizures throughout the country, there has been a corresponding escalation of problems confronting state and local agencies that are called to the scene of these laboratories. \\
% \small 
\textit{Original Target Word:} clandestine\\
\textit{Avg. Score:} 1.89\\ \hline
\end{tabular}
\caption{
The multi-sentence context dataset, ~\cite{kapelner2018predicting} consists of passages collected from the DictionarySquared database.
They also collected crowdsourced ratings of the passage's informativeness level ($(1, 4)$) as the target word is presented in the passage. 
}
\label{tab:multi_sentence_context_ex}
\end{table}

In contextual word learning, the meaning of a target word can be also determined from information in multiple surrounding sentences. 
To test the generalizability of our models to the multi-sentence scenario, we used an existing dataset from the only previous study, to our knowledge, on contextual informativeness~\citep{kapelner2018predicting} (Table \ref{tab:multi_sentence_context_ex}).
Those authors selected 933 words for advanced exams such as the ACT, SAT, and GRE. 
Based on these target words, they collected 67,833 contexts from the DictionarySquared database. On average, each target word had 72.7 ($\sigma^2$=20.7) contexts. 
They categorized target words into 10 difficulty levels, and these levels were not correlated with annotated informative scores.

The multi-sentence context dataset is substantially different from the single-sentence context dataset.
%: not only the length of contexts per target word, but also the size and source of contexts, the range of the target words, and the annotation process.
First, the multi-sentence dataset contains over 67k passages selected from the existing database ($\mu= 81$ words, $\sigma^2 = 42$). 
Second, each multi-sentence context contains one of 933 unique target words, which were selected to range across difficulty levels. 
% This dataset was also designed for contextual vocabulary learning. 
Third, in contrast to our BWS procedure, crowdworkers for the dataset annotated the informativeness of context passages (with target word included) using a four-point Likert scale (roughly corresponding to the four categories in~\cite{beck1983vocabulary}). 
% They also included the target word in the passage.

%%%%%%%%%%%%%%%%%%%%%%%%%%%%%%%%%%%%%%%%%%%%%%%%%%%%%%%%%%%%%%%%%%%%%%%%%%%%%%%%%%%%%%%%%%%%%%%%%%%%%%%%%%%%%%%%%%%

% \section{Experiment 1: Predicting Contextual Informativeness}
\section{Experiment 1: Predicting Informativeness}
\label{sec:exp1}
Using the attention-based contextual informativeness model from Section \ref{sec:contextual_informativess_model}, we conducted four experiments to examine how the model effectively predicts the contextual informativeness of instructional contexts (Experiments 1 and 2) and its application to more efficient training of machine learning models (Experiments 3 and 4). 
In Experiment 1, we analyzed the model's prediction performance in two educational datasets: single- and multi-sentence contexts.

\subsection{Baseline Models and Metrics}
\label{sec:exp1__baselines}
% - simple baselines: dummy guessing, BoW, sentence length -- justify why these can be a simple, but good baselines to compare
% - Kapelner et al: Random forest
% - vanilla BERT
% - move ELMo based models/results to Appendix
\subsubsection{Baselines}
We used multiple baseline models for the analysis.
For simple baselines, we used a dummy model (\texttt{Base:Avg}) that always predicts the average informativeness score from a cross-validation fold; a linear regression model (\texttt{Base:Length}) based on sentence length; and a ridge regression model (\texttt{Base:BoW}) using the co-occurrence information of context words in a bag-of-word style.
These simple baseline models represent a prediction \textit{independent} of the target word. In other words, the models reflect the informativeness of the sentence for all words in a context, regardless of which is the target and ignore the fact that some words may be more or less constrained in their meaning. 

We also used pre-trained models for comparison. We used ELMo (\texttt{Base:ELMo})~\citep{peters2018deep} and BERT (\texttt{Base:BERT})~\citep{devlin2019bert} to predict the contextual informativeness scores only with a regression head (i.e., without the attention block from Figure \ref{fig:mod_structure}). These pre-trained baselines are expected to perform better than the simple baselines, especially on multi-sentence contexts where input texts are longer and more complex.

For the multi-sentence context dataset, we additionally included the random forest model (\texttt{Base:RF\_Lex}) from~\cite{kapelner2018predicting}. Thanks to data provided by those authors, our implementation of the random forest model was able to replicate their results, with almost identical $R^2$ scores (e.g., 0.179 vs. 0.177), using lexical features of contexts, such as the top 10 words that include synonymous words from the target, top 10 context words that frequently collocate with the target words, frequency of the target word, context words' politeness, age of acquisition, and meaningfulness of context words.
% \footnote{We could not regenerate all of the 600+ lexical features for the single-sentence dataset since access to feature engineering resource presented in~\cite{kapelner2018predicting} was limited. }.

\subsubsection{Metrics}
To measure the prediction performance of our model, we used two metrics.
First, RMSE shows how the model's prediction scores diverge from the true scores. 
Second, for ROCAUC, we set specific thresholds to investigate how the model performed in classifying binary labels for high (e.g., top 20\% or 50\%) or low (e.g., bottom 20\%) informative sentences. 
The high-precision setting resembles the goal of the previous study ~\citep{kapelner2018predicting}.
Also, per Section~\ref{sec:related__context_literacy}, selecting a range of context informativeness levels can be important in learning applications to develop curricula based on the contextual informative scores. 
All reported results are based on 10-fold cross-validation. Each fold was randomly selected while stratified by the target word, to ensure the model did not see the sentence with the same target word during the training process.

\subsection{Results}

\label{sec:exp1__res}
\begin{figure*}[t]
\centering
\includegraphics[width=0.95\linewidth]{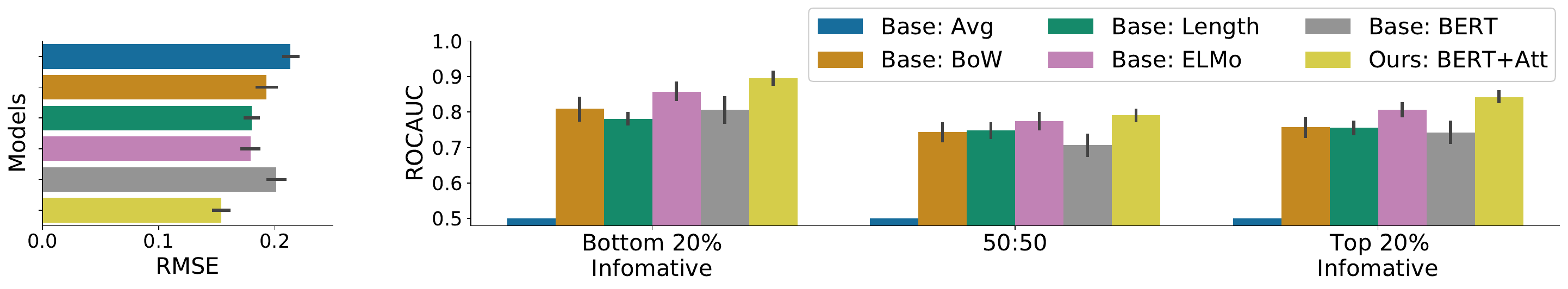}
\caption{
RMSE (lower is better) and binary classification results (higher is better) with the single-sentence context dataset. 
Among the simple baseline models, the sentence-length baseline (\texttt{Base:Length}) and co-occurrence baseline (\texttt{Base:BoW}) models showed similar results.
For the pre-trained baselines, the ELMo baseline (\texttt{Base:ELMo}) showed better classification results than other simple baseline models, while the BERT baseline (\texttt{Base:BERT}) did not. 
However, adding the attention block to BERT (\texttt{Ours:BERT+Att}) significantly improved the prediction performance from all baseline models.
}
\label{fig:pred_res_single}
\end{figure*}

\begin{figure*}[t]
\centering
\includegraphics[width=0.95\linewidth]{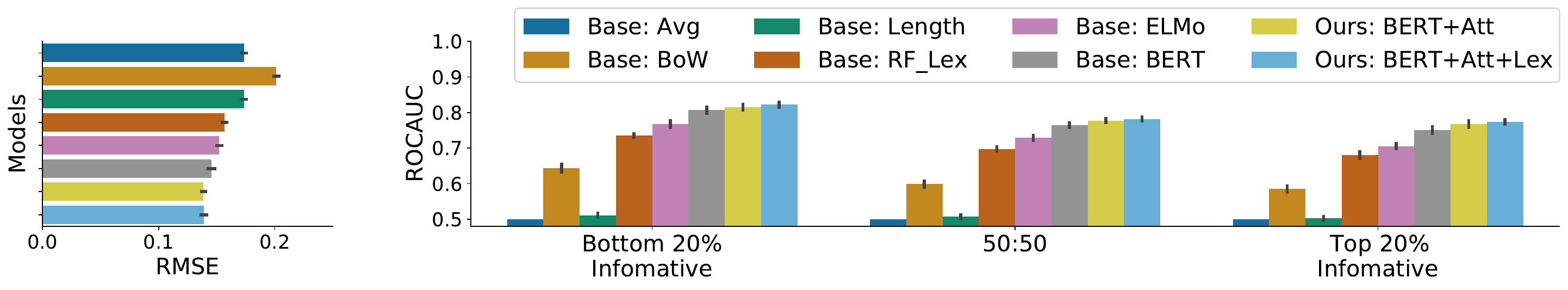}
\caption{
RMSE and binary classification results with the multi-sentence context dataset~\citep{kapelner2018predicting}. 
For the multi-sentence contexts, the pre-trained baseline models showed significantly better performance than the simple baseline models. 
The hybrid model (\texttt{Ours:BERT+Att+Lex}) that combined our attention features with lexical features from~\cite{kapelner2018predicting} achieved the best prediction performance overall.  
}
\label{fig:pred_res_multi}
\end{figure*}

For both single- and multi-sentence contexts, we found that our model with attention block performed better than all baseline models. We also found that our model was able to effectively predict the informativeness scores of our single-sentence contexts when trained with the multi-sentence corpus from~\cite{kapelner2018predicting}, showing that our model can capture salient aspects of contextual informativeness across very different datasets. More details can be found in Appendices~\ref{app:exp1_results_single} and ~\ref{app:exp1_results_multi}.

\subsubsection{Single-Sentence Contexts}
\label{sec:exp1__res_single}

% - prediction results: single/multi-sentence contexts
% \daj{Is there some way to doing identifying the subset of data that are ``harder'' contexts for the baselines and then show results on these separately? }

We found that adding the attention layer significantly improved the prediction. In most cases, our BERT-based model with attention block (\texttt{BERT+Att}) performed better than the baselines in RMSE and ROCAUC metrics (Figure~\ref{fig:pred_res_single}). 
% 95\% confidence intervals (CIs) of 50:50 classification were 0.736-0.775 (\texttt{Base:Length}), 0.784-0.827 (\texttt{Base:ELMo}), and 0.824-0.860 (\texttt{BERT+Att}).
Adding the attention block to ELMo (\texttt{ELMo+Att}) only showed marginal improvements.
% , 0.787-0.834 (\texttt{ELMo+Att}), 

% With single-sentence contexts, 
The performance of the sentence-length baseline (\texttt{Base:Length}) showed that raw word count provides much information about contextual informativeness. 
This is expected for short or single-sentence contexts where more context words may correlate with contextual informativeness scores. The co-occurrence baseline model (\texttt{Base:BoW}) showed a similar level of performance.

For pre-trained baselines, the ELMo baseline (\texttt{Base:ELMo}) showed better classification results than the simple baseline models. 
The BERT baseline (\texttt{Base:BERT}) shows surprisingly low performance. We think this may be caused by the mismatch between the large number of parameters in BERT and the relatively small number of sentences in the single-sentences dataset. 
% This is a characteristic of the single-sentence context dataset, and also indicates that improving the performance of predicting contextual informativeness can be challenging for this dataset. 
% More details about the experiment results can be found in Section~\ref{app:exp1_results_single}.

% Adding the attention block to the model introduced either slight (ELMo-based) or significant (BERT-based) improvement (Appendix~\ref{app:pred_dscovar}). 
% Additional results can be found in Appendix~\ref{app:pred_dscovar}.
% Adding the attention block (\texttt{+Att}) to the model also introduced significant improvement for the BERT-based model. The ELMo-based model showed similar but marginal improvements. 

% From the results, we can conclude that adding the attention block can improve the informativeness prediction performance with single-sentence contexts. 
% Similar results were observed for \textit{semantic density} scores. However, the performance gain from the attention block seems less significant. 

\subsubsection{Multi-Sentence Contexts}
\label{sec:exp1__res_multi}

% For multi-sentence contexts, t
Adding the attention block (\texttt{BERT+Att}) provided marginal gains over the BERT baseline (\texttt{Base:BERT}). %(e.g., 0.770-0.785 (\texttt{BERT+Att})). 
The complementary model (\texttt{BERT+Att+Lex}), which concatenated attention-weighted context vectors with lexical features, provided significantly better performance than all baseline models (Figure \ref{fig:pred_res_multi}). % (0.775-0.788 (\texttt{BERT+Att+Lex})). 
% More detailed results on predicting multi-sentence contexts can be found in Appendix~\ref{app:exp1_results_multi}. 

The sentence-length baseline (\texttt{Base:Length}) showed near-random performance in classification, since multi-sentence contexts were long enough that the amount of informativeness of each context was less correlated to the number of words. The co-occurrence baseline model (\texttt{Base:BoW}) showed significantly better performance than \texttt{Base:Length}, but not better than the pre-trained baselines. All ELMo- and BERT-based baseline models performed significantly better than other baseline models. 
% 95\% CIs in 50:50 classification were 0.691-0.705 (\texttt{Base:RF\_Lex}), 0.721-0.737 (\texttt{Base:ELMo}), and 0.757-0.772 (\textit{Base:BERT}).
%  0.720-0.734 (\texttt{ELMo+Att}),
More details about the experiment results can be found in Sections~\ref{app:exp1_results_single} and \ref{app:exp1_results_multi}.

\subsubsection{Cross-Training}
\label{sec:exp1__res_cross}

\begin{table}[t]
\centering
% \begin{tabular}{p{1in} p{0.4in} p{0.4in} p{0.4in} } \hline
\begin{tabular}{r l l l} \hline
ROCAUC      & $\downarrow$ 20\% Info     & 50:50    & $\uparrow$ 20\% Info \\ \hline
Multi $\rightarrow$ Single    & 0.784 & 0.752 & 0.715 \\
Single $\rightarrow$ Multi    & 0.502 & 0.497 & 0.492 \\\hline
\end{tabular}
\caption{
ROCAUC results for cross prediction between datasets. 
\texttt{BERT+Att} model trained on the multi-sentence dataset were effective at predicting contextual informativeness scores for the single-sentence context dataset. % (the BERT-based model giving better results). 
However, less effective prediction results were observed in the other direction. 
}
\label{tab:pred_res_cross}
\end{table}

We also tested the generalizability of our model by cross-training. 
Table \ref{tab:pred_res_cross} demonstrates that the model trained with the multi-sentence context dataset was effective at predicting the contextual informativeness scores for the single-sentence context dataset (\texttt{Multi $\rightarrow$ Single}), showing that our model is capturing some essential aspects of contextual informativeness.
% In this scenario, our \texttt{BERT+Att} model performed better than random chance (0.50). 
However, the reverse scenario (\texttt{Single $\rightarrow$ Multi}), where the same model structure was trained with the much smaller single-sentence context dataset, was not effective in predicting multi-sentence context scores. We believe this is likely due to the greatly reduced number of training examples in the single-sentence dataset as well as the lack of variety of contexts in each very short, single-sentence example.
%The weaker performance in this transfer learning setting suggests that models are learning context-type-specific cues of contextual informativeness, rather than general strategies, implying that general contextual informativeness models should see a variety of contexts to perform well.

%The previous experiments showed that adding the attention block improved prediction performance for single- and multi-sentence contexts.
% We tested the generalization ability of our model by cross-training: 
% when the model was trained on one dataset (e.g., multi-sentence), evaluate if the model could effectively predict the contextual informativeness scores from another (e.g., single-sentence).
% evaluating whether, when trained on one dataset (e.g., single-sentence) the model could effectively predict the contextual informativeness scores for a different dataset (e.g., multi-sentence).

%%%%%%%%%%%%%%%%%%%%%%%%%%%%%%%%%%%%%%%%%%%%%%%%%%%%%%%%%%%%%%%%%%%%%%%%%%%%%%%%%%%%%%%%%%%%%%%%%%%%%%%%%%%%%%%%%%%

\section{Experiment 2: Evaluating Attention Weights}
\label{sec:exp2}
In Experiment 1, we showed that our attention-based model effectively predicts contextual informativeness scores for two different datasets. 
In Experiment 2, we examined how the model's attention mechanism provides interpretable details on informative contexts, by identifying contextual cues that facilitate a more precise inference of the meaning of the target word.

\subsection{Dataset: Attention Evaluation}
\label{sec:exp2__dataset}
% - EVALution dataset
%     - better baseline than random score
%     - compare with vanilla BERT model: last-layer's attention of target word
To evaluate our model's attention output across different types of contextual relationships, we used the EVALution dataset~\citep{santus2015evalution}. This dataset was originally designed for evaluating whether word embedding models capture nine different semantic relations between word pairs. In our study, we used the dataset to quantify how well the contextual informativeness model captures relevant words as important contextual information  (Section \ref{sec:exp2__res_evalution}). The dataset includes over 1800 unique words and seven thousand word-pairs, used in automatically generated example sentences.

Each sentence of the EVALution dataset mirrors lexico-syntactic patterns expressing a particular semantic relationship~\citep{hearst1992automatic,pantel2006espresso,snow2005learning}, so a contextual cue to the meaning of a target word is easily identifiable. Although formulaic, this regular structure allows us to control contextual confounds and to test which types of relational information are identified by the model.
% We can identify the textual words that are more important to infer the meaning of the target word, namely the pair word and relational cue words. 
% ``An account is a type of record that stores transactions.''

For example, in the sentence ``An account is a type of record,''  the meaning of the target word  ``account'' is informed by a single contextual \textit{pair word} ``record'', and a \textit{relational cue} word ``type'' that says how the pair word relates to the target semantically. 
% We expect that the attention mechanism to capture an interpretable explanation of informativeness, by putting more weights on the pair and relational cue words. 
If the attention mechanism reflects an interpretable explanation of informativeness, we expect to observe more weights on the pair word and relational cue word in the context (e.g., ``record'' and ``type'').
In Section~\ref{sec:exp2__res_qualitative}, we also included some additional hand-picked examples to show how the additional attention layer captures contextual informativeness from different sentences, with respect to the masked target word.

\subsection{Baseline Models and Metrics}
For the analysis, we used two types of baselines: a random rank assignment for the pair or relational cue words; and the average of attention head scores from selected layers of an off-the-shelf BERT model.
For the BERT-based baselines, we calculated the average of attention weights for context words across multi-heads for each layer, with respect to the target word. Each layer of BERT may represent different types of semantic relationships between context words~\citep{clark2019does}. We included the results from the selective layers (first (\texttt{BERT\_Early}), sixth (\texttt{BERT\_Mid}), and twelfth (\texttt{BERT\_Later}) layers) of vanilla BERT layers to compare with the outcomes of our model's attention block. 
% from the additional attention block.

This experiment was to examine whether models rank more salient contextual cues highly. To do this, we used the normalized rankings of the \textit{pair} or \textit{relational cue} words' attention weights from each sentence ($1-\frac{rank(x)-1}{N-1}$; $N$= the number of words in a sentence).

\subsection{Results}

\subsubsection{Rank Context Words}
\label{sec:exp2__res_evalution}

\begin{figure*}[ht!]
\centering
\includegraphics[width=0.99\linewidth]{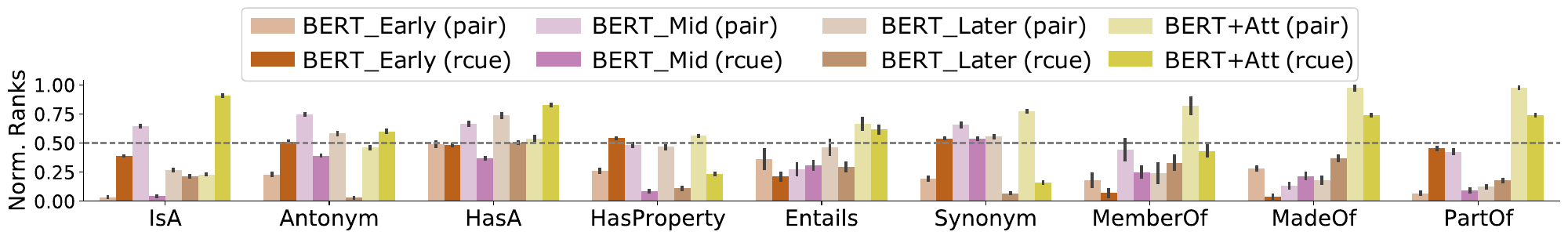}
\caption{
Comparing the normalized rank scores (higher is better) on the EVALution dataset~\citep{santus2015evalution}. Our BERT-based model with the attention block (rightmost 2 scores per group, \texttt{BERT+Att}) was more successful at identifying important context words, capturing both \textit{pair} and \textit{rcue} contexts better than the random baseline in 4 of 9 relations. 
For comparison, the off-the-shelf BERT layers consistently did not perform well: 
For example, only 1 relation from each 6th layer (\texttt{BERT\_Mid}; \textsc{Synonym}) and 12th layer (\texttt{BERT\_Later}; \textsc{HasA}) captured both \textit{pair} and \textit{rcue} contexts better than the random baseline (dotted line; normalized rank score of 0.5). 
% Adding the attention block to the model introduced significantly better improvements on capturing the relational cue word when the model also fine-tuned the pre-trained BERT layer. 
% The no-tuned model performed better in capturing the pair word in selective relations. 
}
\label{fig:exp2_att_evaluation_scores}
\end{figure*}

% - more qualitative interpretations? e.g., single \& multi sentence examples
Figure~\ref{fig:exp2_att_evaluation_scores} shows that our attention-based model structure (\texttt{BERT+Att}; trained with the multi-sentence context dataset) effectively guided the vanilla BERT model to predict contextual informativeness with respect to the target word.
%  compares the performance of our contextual informativeness model to the baseline models on the EVALution dataset~\citep{santus2015evalution}. 
% in instructional settings, while the off-the-shelf BERT model only occasionally performed better than the random baseline.
For example, our \texttt{BERT+Att} model successfully captured both the pair and relational cue words in 4 of 9 relations, 7 of 9 cases for pair words, and 6 of 9 cases for relational cue words. 
Our model did not perform well at ranking the pair word in \textsc{IsA} and \textsc{Antonym} relationships, despite these pair words providing the most evidence of meaning.
However, the model was able to correctly rank meronyms (e.g., \textsc{MadeOf}, \textsc{PartOf} , and \textsc{MemberOf}).
% as having high semantic information.
The \texttt{ELMo+Att} model captured the pair words better. However, it did worse with the relational cue words. More details can be found in Appendix~\ref{app:exp2_evalution}. 

In comparison, off-the-shelf BERT layers only captured both the pair and relational cue words only in very limited relations (e.g., \texttt{BERT\_Mid} with \textit{Synonym}).
The first layer of the baseline BERT model (\texttt{BERT\_Early}) only captured the relational cue words in 3 of 9 relations (\textsc{Antonym}, \textsc{HasProperty}, \textsc{Synonym}).
The sixth layer (\texttt{BERT\_Mid}) captured the pair words in 4 of 9 relations (\textsc{IsA}, \textsc{Antonym}, \textsc{HasA}, \textsc{Synonym}) and the relational cue words in 1 of 9 relation (e.g., \textit{Synonym}).
The last layer (\texttt{BERT\_Later}) captured the pair words better than the random baseline in 3 of 9 relations (e.g., \textsc{Antonym}, \textsc{HasA}, and \textsc{Synonym}), and the relational cue words in 1 of 9 relation (e.g., \textit{HasA}). 
% More detailed results can be found in Appendix~\ref{app:exp2_evalution}.

\subsubsection{Qualitative Evaluation}
\label{sec:exp2__res_qualitative}

\begin{figure}[t]
\centering
\includegraphics[width=\linewidth, height=1.1in]{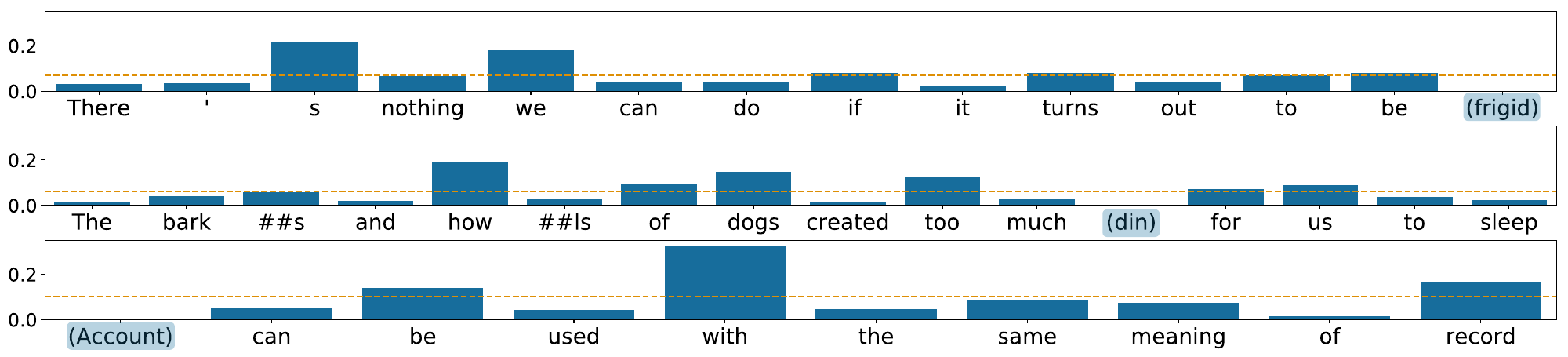}
\caption{
Example attention weight distributions from our \texttt{BERT+Att} model. The first sentence is low-informative, and the second and third sentences are high-informative. Our model assigns high attention weights to informative terms
% (e.g., context words) 
that are contextually relevant to the target words in the high-informative sentences. 
%For the second sentence, 
% The first two sentences are the low and high informative sentence examples from Table \ref{tab:ex_sentences}. 
%weights for relevant content words (e.g., din: `howls'
% `sleep' -- interpretation for incorrect results
%) are higher than the average weights (dotted line). 
%The last sentence is a synonym-relation example from~\cite{santus2015evalution}. 
%The model put more attention weights on the pair word (\textit{record}).
% the relational cue word (\textit{same}). -- interpretation for incorrect results
The dotted line marks the average weight value for each sentence.
Each target word (highlighted) was masked for the model. 
}
\label{fig:attn_example}
\end{figure}

Figure \ref{fig:attn_example} shows more details on how our model learned a relationship between the target word and context words with example sentences. 
The first two sentences are drawn from our single-sentence context dataset~\cite{frishkoff2016dynamic}.
For the (first) low-informative sentence, the model put more weight on function tokens (e.g., \textit{'s, we}), suggesting that the sentence lacked sufficient content words to constrain the meaning. 
However, for the (second) high informative sentence, the model tended to put more weight on individual content words (e.g., \textit{howls, dogs, too}) that help infer the meaning of the target word. 
The last sentence shows the output for a synonym-relation sentence~\citep{santus2015evalution}. The model's attention activation successfully highlighted the pair word (\textit{record}) or a preposition that can imply the target-pair relationship (\textit{with}), but not the relational cue word (\textit{same}).
Further analysis is needed to examine why some context cues were not highlighted, and how methods for fine-grained interpretation of contextual informativeness could help curriculum instruction or improve prediction performance.

%%%%%%%%%%%%%%%%%%%%%%%%%%%%%%%%%%%%%%%%%%%%%%%%%%%%%%%%%%%%%%%%%%%%%%%%%%%%%%%%%%%%%%%%%%%%%%%%%%%%%%%%%%%%%%%%%%%
\section{Experiment 3: Curriculum Learning for Batch Learning Models}
\label{sec:exp3}
In Experiments 1 and 2, we introduced our attention-based model and showed the model can predict the amount of contextual informativeness in single- and multi-sentence contexts. We also confirmed that our model effectively identifies more important context words with respect to the target word. 

In the next experiments, we explored whether our contextual informativeness model, originally developed to help students, could also be used to improve curriculum learning for a reading-like machine learning task. 
For Experiment 3, we evaluated word embedding models which, analogous to human readers, rely on multiple exposures to a word's context to form their word representation. We used our contextual informativeness prediction model to control the order and proportion of informative examples that made up a word embedding's training curriculum.

\subsection{Models, Dataset, and Task}
\label{sec:exp3__model_dataset_task}

\subsubsection{Batch Learning Models}
\label{sec:exp3__model}
% - Word2Vec / Fast-text models
In Experiment 3, we examined the effect of different training curricula based on contextual informativeness scores in a batch learning scenario, where the word embedding model learned new target words with a large number of sentences at once. 
For the experiment, we used two word embedding models: \textit{Word2Vec} and \textit{FastText}. 
We chose Word2Vec~\citep{mikolov2013distributed} since it is widely used and has been a strong traditional baseline in many NLP tasks.
FastText~\citep{bojanowski2017enriching} is an extension of Word2Vec that incorporates a character-level encoding. It is known to handle rare words better by representing out-of-vocabulary words through n-gram information. 

% we tested different learning scenarios, \textit{batch learning} scenario and \textit{few-shot learning} scenario. 
% We evaluated how each curriculum based on contextual informativeness scores can affect word embedding models' performance across different numbers of target sentences used per target word. 
% The first experiment tested \textit{Word2Vec} and \textit{FastText} models.

\subsubsection{Curricula Based on Contextual Informativeness}
\label{sec:exp3__dataset_curriculum}

% - Corpus
To build curricula for word embedding models, we used \textit{ukWaC} corpus \citep{ferraresi2008introducing}. ukWaC is a corpus of British English collected from web pages with the \textit{.uk} domain, using medium-frequency words from the British National Corpus as seed words. Since the corpus is collected from a broad set of web pages, its sentences contain different types of writings with various contextual informativeness levels. 

% - heuristics for curriculum building
To carefully control the learning materials and compare outcomes between curricula, we chose sentences from the corpus as follows.
First, we divided sentences into \textit{non-target sentences} and \textit{target sentences}. 
\textit{Non-target sentences} did not contain any target word from the three tasks explained in the following Section~\ref{sec:exp3__task} (8.1M sentences, 2.4M unique tokens, 133.7M tokens). 
% 8,057,490 sentences, 2,363,595 unique tokens, 133,744,553 total tokens 
Each \textit{target sentence} contained only one the target word. 
All sentences containing multiple target words were removed from the analysis. 
During the curriculum building process, target sentences were scored with the contextual informativeness model (\texttt{BERT+ATT} model trained with multi-sentence contexts) and used for developing curricula per heuristic strategies~(Table \ref{tab:curriculum_list}).

Second, we selected sentences that were 10--30 words long, which is similar to the length of average English sentences (15-20 words)~\citep{cutts2013oxford}.
This criterion filtered out sentences that are too short or too long. 
It also helped to control for potential correlation between sentence length and contextual informativeness scores (e.g., our \texttt{Base:Length} baseline model in Section \ref{sec:exp1}), and kept the number of words between curricula relatively similar. 
Details on the distribution of informativeness scores and relationship with sentence length are in Appendix \ref{app:exp3__scores_and_sentences}.

% From the preliminary analysis,
% (Sections \ref{sec:preliminary_analysis} and \ref{sec:res_exp1_predict_infomativeness}), 
% we observed that the length of the sentence is weakly related to the level of informativeness, and can be used as a baseline for predicting the contextual informativeness scores. 

Third, we further filtered out the target words with a sufficient number of acceptable training sentences in the corpus.
We sampled 512 sentences per target word. Each target sentence contained the target word only once.
If a target word had less than 512 training sentences, we excluded the target word from the analysis. 
As a result, our analysis covered 94.16\% (SimLex-999; 968 of 1028 unique target words), 80.89\% (SimVerb-3500; 669 of 827), and 95.65\% (WordSim-355; 418 of 437) of target words of each task. 
% More details of selecting sentences are described in each learning scenario (Sections \ref{sec:learning_scenario_batchlearning} and \ref{sec:learning_scenario_fewshotlearning}). 
% simlex: 968/1028
% wordsim: 669/827
% simverb: 418/437
% \section{Learning Scenarios}
% \label{sec:learning_scenarios}

\begin{table}[t]
\centering
% \small
% \resizebox{0.48\textwidth}{!}{
\begin{tabular}{p{1in} p{2.2in}} \hline
% \begin{tabular}{l l} \hline
\textbf{Curriculum}     & \textbf{Description} \\ \hline
Low Informative      & Selecting bottom $k$ informative sentences per target word.\\
High Informative     & Selecting top $k$ informative sentences per target word.\\
Rand. Select   & Random $k$ sentences per target word.\\
Rand. Non-Low  & Random $k$ sentences per target word from the top half informative sentences (256).\\
Rand. Non-High & Random $k$ sentences per target word from the bottom half informative sentences (256).\\ \hline
\end{tabular}
% }
\caption{
Curriculum heuristics based on contextual informativeness.
}
\label{tab:curriculum_list}
\end{table}

With these criteria, we developed the training curriculum by selecting $k$ ($k=2^i, i\in\{1, ..., 9\}$) \textit{target sentences} for each target word, by using five simple heuristics (Table \ref{tab:curriculum_list}). 
Note that for the non-low or non-high randomly selected sentence curricula, the maximum number of sentences per target word was 256. Essentially, this was equal to the 256 most high- or low-informative sentences. For these curricula, we did not include the results for 512 sentences per target word.

\subsubsection{Evaluation Tasks}
\label{sec:exp3__task}
\begin{figure}[t]
    \centering
    \includegraphics[width=0.8\linewidth]{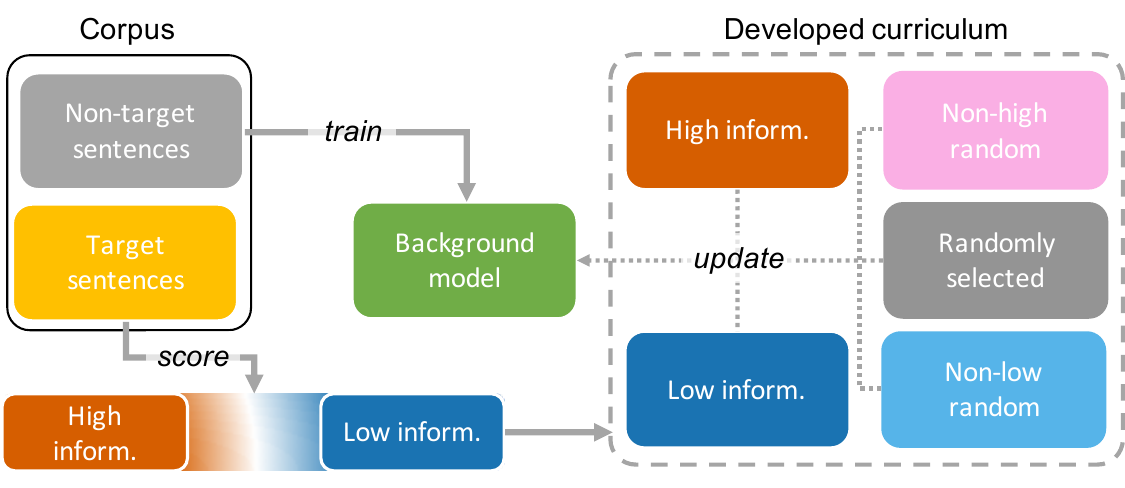}
    \caption{
    An illustration of curriculum developing process. 
    First, we separated sentences into \textit{non-target} and \textit{target sentences}, based on if the sentence contains any target word from the semantic similarity tasks. 
    Second, \textit{non-target sentences} were used to train the background model. \textit{Target sentences} were scored by the contextual informativeness model.  
    % (e.g., the BERT-based model trained with multi-sentence contexts).
    Third, different curriculum heuristics were developed to update the background model with \textit{target sentences}. 
    }
    \label{fig:curriculum_setting}
\end{figure}

% After the task was firstly introduced, it was criticized for not clearly distinguishing the similarity and relatedness for the annotation process. In our experiment, we additionally included the distinction from~\citep{agirre2009study}, which separated the word pair scores into similarity and relatedness score sets. 

% - model inherent evaluations (e.g., median rank, perplexity?)
% - Semantic similarity datasets
% For Experiment 3, we compared how different curricula based on contextual informativeness scores can bring different results for updating the same pre-trained background word embedding model. 
% We also compared how curriculum building strategies interact with different sentence sizes per target word.  

% First, we trained the background model with \textit{non-target sentences}. This simulates a model with existing knowledge of English words excluding the target words.
% Target words of each semantic similarity task were treated as unseen words to be learned by word embedding models from \textit{target sentences}. 
%
% Second, 
% For the non-low or non-high informative sentence curricula, we filtered 256 low or high informative sentences respectively (i.e., half of the target sentences per target word) to control the range of sampled sentences. After this, we sampled $k$ sentences for each curriculum.}. 
We used three semantic similarity tasks to evaluate word embedding models in batch learning settings. 
First, \textit{SimLex-999}~\citep{hill2015simlex} included 999 pairs of nouns, verbs, or adjectives. 
% It also distinguished more associated word pairs. 
Second, as the name suggests, \textit{SimVerb-3500}~\citep{gerz2016simverb} included verb pairs. It used the same guidelines as SimLex-999 for collecting human annotations. 
Third, \textit{WordSim-353}~\citep{finkelstein2002placing} included scores for noun pairs. 
These tasks used human annotations on semantic similarity between word pairs as their gold standard. 

To compare the effects of different curricula, we first trained a word embedding model (e.g., \textit{Word2Vec} or \textit{FastText}) with \textit{non-target sentences} as a background knowledge model. This simulated the model's prior knowledge on other words. 
Then we updated the background model of each task by varying the curriculum heuristics and the number of \textit{target sentences} per target word (Figure \ref{fig:curriculum_setting}), letting the model to learn about target words.  
Lastly, we calculated the cosine similarity scores for word pair vectors and compared the Spearman's rank correlation to the human-annotated scores.
% To evaluate the learning performance for each semantic similarity task, we used Spearman's rank coefficient between cosine similarity scores of word pair vectors and human-annotated scores. 
% Later, the non-target sentences were used to train the background model, which simulates the model's prior knowledge on other words. 
% Then the background model was updated with a curriculum consisting of target sentences, and acquired knowledge on target words, 

\subsection{Results}
% - explain evaluation metrics: median rank/correlation
\label{sec:exp3__results}

% \subsubsection{SimLex-999, SimVerb-3500, and WordSim-353}
\subsubsection{Word Pair Similarity Tasks}

\label{sec:exp3__results_all_tasks}
\begin{figure*}[t]
\centering
% \hspace*{-0.61in}
\includegraphics[width=1.0\linewidth]{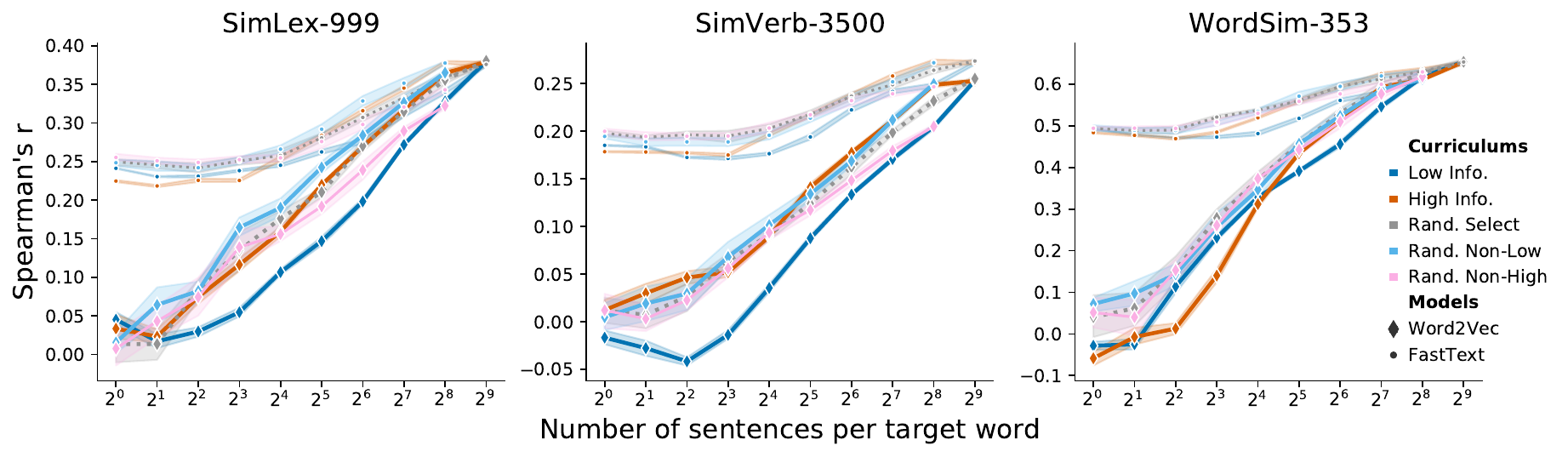}
\caption{
Batch learning results on Word2Vec\textsuperscript{\sdiam} and FastText\textsuperscript{$\bullet$} models with SimLex-999, SimVerb-3500, and WordSim-353 tasks. Shades represent the 95\% CI. 
The results show that the contextual informativeness score can successfully distinguish less helpful sentences for the batch learning models. 
With fewer training sentences, the FastText models are significantly better than Word2Vec models. 
Notice that y-axis scales are different between the tasks.
% For example, the low informative sentence models performed worst in most cases. 
% but both model types' performance converge as the number of sentences per target word increases.
}
\label{fig:curriculum_result}
\end{figure*}

\begin{figure*}[t]
\centering
\includegraphics[width=1.0\linewidth]{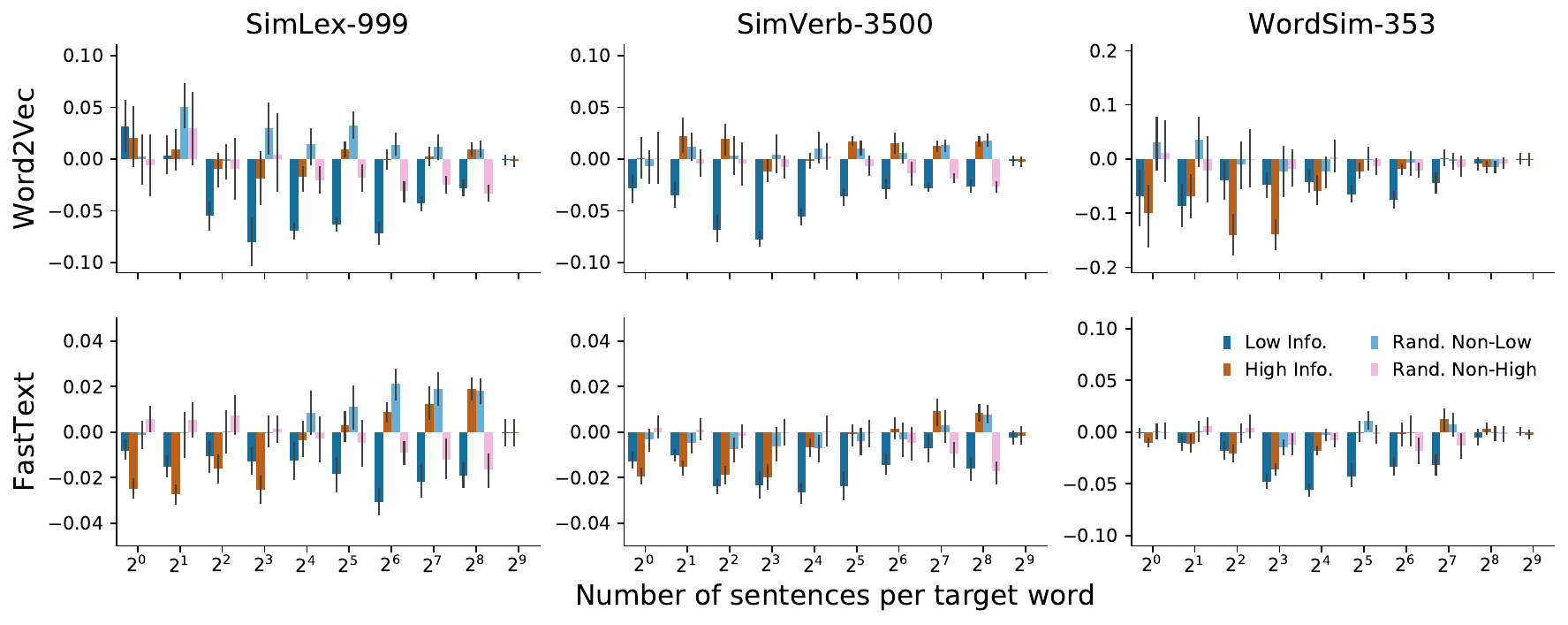}

\caption{
Performance gain from the random curriculum models. Vertical lines represent the 95\% CI. 
In most cases, low-informative sentence models (\texttt{Low Info.}) performed worse than random baselines. 
Filtering out low-informative sentences (\texttt{Rand. Non-Low}) increased embedding models' performance in many cases, 
% For SimLex-999 and SimVerb-3500, as the number of sentences per target word increased, the non-low informative sentence models performed significantly better than others, 
while the non-high informative sentence models (\texttt{Rand. Non-High}) often performed worse than the random curriculum models. 
Note that scales are smaller for FastText models but still show consistent trends.
% For WordSim-353 task, there were no meaningful improvements of different curriculum from the random sentence models.
}
\label{fig:curriculum_result_diff}
\end{figure*}

Word2Vec is known to perform at various levels for the similarity tasks. 
The reported Spearman's rank coefficient with human-annotated scores were 0.414 (SimLex-999)~\citep{hill2015simlex}, 0.274 (SimVerb-3500)~\citep{gerz2016simverb}, and 0.655 (WordSim-353)~\citep{hill2015simlex}. 
Although we used a different corpus and parameters, we were able to achieve similar performance with our Word2Vec models. 

Figure~\ref{fig:curriculum_result} demonstrates that our contextual informativeness model successfully distinguished less helpful training sentences from other sentences from the corpus. The low-informative sentence curriculum (\texttt{Low Info.}) provided significantly worse performance. We also observed that the non-low-informative sentence curriculum (\texttt{Rand. Non-Low}) performed better than others. 
% All curricula performance were converged as the size of target sentences reached the maximum ($> 2^9$).

Figure \ref{fig:curriculum_result_diff} shows more clearly that
the non-low informative sentence (\texttt{Rand. Non-Low}) or the high informative sentence models (\texttt{High Info.}) tended to perform better than other curricula. 
The non-high informative sentence (\texttt{Rand. Non-High}) or the low informative sentence models (\texttt{Low Info.}) tended to perform worse than these models. 
These results indicate that contextual informativeness scores may improve the word embedding models' performance by removing less-informative contexts from the training set.
% Figure \ref{fig:curriculum_result_diff} shows more detailed comparison between models from the random sentence models' performance. 

For SimLex-999 and SimVerb-3500 tasks, we noticed that FastText models were significantly better and stable than the Word2Vec models when trained with the smaller numbers of training sentences. 
The results confirm that FastText models are likely a better choice for representing rare words~\citep{bojanowski2017enriching} and as such, the differences between curricula are smaller (e.g., $< 2^4$) for FastText models than for Word2Vec models. 
However, the curricula \textit{without} random sampling (e.g., both \texttt{Low Info.} and \texttt{High Info.}) provided worse performance in smaller training sizes, while the non-low informative sentence models (\texttt{Rand. Non-Low}) showed only marginally better results than the random curricula. This may mean that the balance between diversity and quality of training sentences need to be carefully determined for a more sensitive model like FastText. 
Using all training sentences per target word ($2^9$), the performance of these models eventually converged within a similar score range.

With the WordSim-353 task, we observed mixed results, especially with Word2Vec models. The high-informative sentence models (\texttt{High Info.}) performed better than the low-informative sentence models (\texttt{Low Info.}) when the number of training sentences was $2^5$ or greater per target word. However, with the lower number of sentences per target word (e.g., $< 2^5$), the results from the low informative sentence models were better. 
% , but vice versa in the opposite case. 
FastText models showed relatively consistent trends. The low-informative sentence models (\texttt{Low Info.}) performed worse than others in many cases. However, the high-informative sentence models (\texttt{High Info.})or the non-low informative sentences models (\texttt{Rand. Non-Low}) did not perform better than the randomly selected sentence models (\texttt{Rand. Select}). 
% However, both models did not perform better than the random sentence models or the non-low informative random sentence models. 

\subsubsection{SimLex-999: By Part of Speech and Association Pairs}
\label{app:exp3__results_pos_assoc}
% We also crosschecked that similar results were observed between different parts of speech (POS) and association pairs in the SimLex-999 dataset.
Among the three semantic similarity tasks, SimLex-999 has the most variety in target word attributes. In this section, we further analyzed the results from the SimLex-999 task by the target word pairs' part of speech (POS) (222 verb, 666 noun, and 111 adjective pairs) and association (333 associated and 666 non-associated pairs). 

\begin{figure*}[t]
\centering
% \hspace*{-0.61in}
\includegraphics[width=1.0\linewidth]{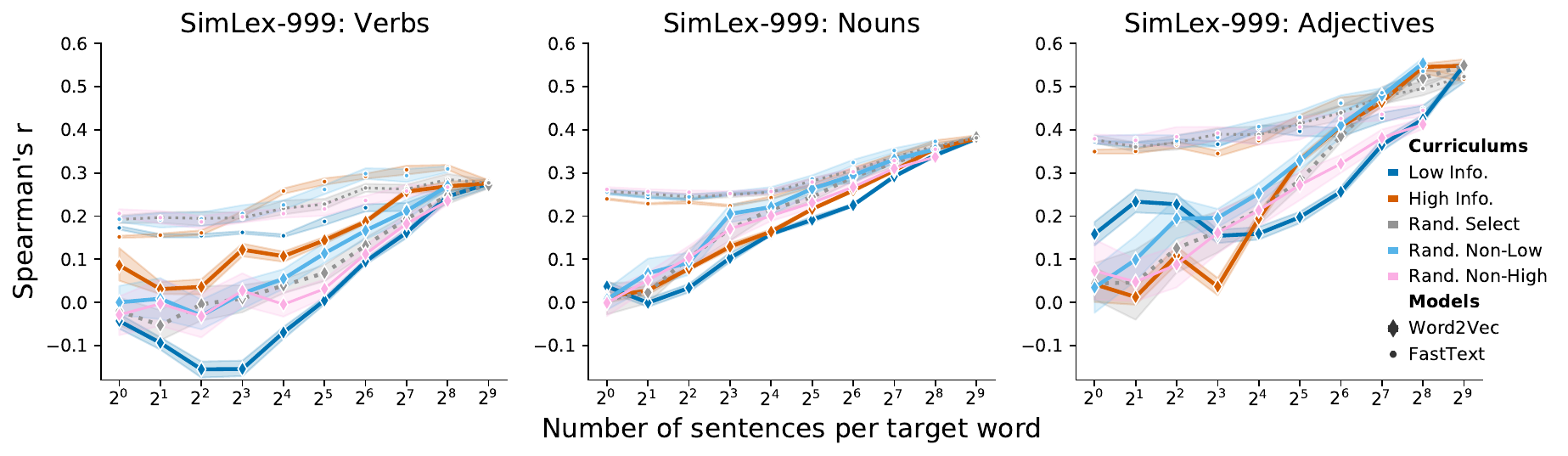}
\caption{
Batch learning results on Word2Vec\textsuperscript{\sdiam} and FastText\textsuperscript{$\bullet$} with different POS of SimLex-999 target words. 
In most sentences per target word condition, filtering out low-informative sentences provided significantly better performance than the model trained only with low-informative sentences (\texttt{Low Info.}).
For verbs, the high-informative sentence models (\texttt{High Info.}) showed significantly better performance than the randomly selected sentence models. 
For nouns, the non-low informative sentence models (\texttt{Rand. Non-Low}) performed significantly better than models trained on random sentences. 
}
    \label{fig:curriculum_result_simlex_pos}
\end{figure*}

\paragraph{By Part of Speech}
\label{app:exp3__results_pos}
The results by different target word POS were similar to the previous results from Simverb-3500 and WordSim-353 tasks (Figure~\ref{fig:curriculum_result_simlex_pos}). 
FastText models still showed more stable and better results, especially with the lower number of training sentences per target word. 

In many cases, the non-low informative sentence models (\texttt{Rand. Non-Low})showed better performance than other curricula models. However, there were some differences between target words' POS. 
With the verb pairs, the high-informative sentence models (\texttt{High Info.}) performed significantly better than the low-informative sentence models (\texttt{Low Info.}), and the high-informative models performed significantly better than the randomly selected sentence models (\texttt{Rand. Select}). 
Similar to the SimVerb-3500 results, a large dip (e.g., $2^2$--$2^3$ sentences per target word) was also observed for Word2Vec models with low-informative sentences (\texttt{Low Info.}). This confirms that the Word2Vec model can be unstable with a smaller number of sentences per target word, especially with verb target words and low-informative sentences. 

For noun pairs, the non-low informative random sentence models (\texttt{Rand. Non-Low}) performed consistently better than other models. With Word2Vec models, low-informative sentence models (\texttt{Low Info.}) performed consistently worse than others, while FastText models with high-informative sentences (\texttt{High Info.}) did not perform well around the $2^5$ sentences range. 
Along with WordSim-353 results, this is evidence that the curriculum effect can vary with the specific noun target.

Lastly, for adjective pairs, we observed unusual examples of early learning in the $2^0$--$2^3$ sentence per target word range from the low-informative sentence (\texttt{Low Info.}) Word2Vec models.
This may due to the small number of target words that are adjectives. This early advantage did not link to as much continuous improvement as the model trained with more sentences. 
% Similar to the WordSim-353 results, we did not see the performance improvement from the randomly selected sentence models. 

\begin{figure*}[t]
\centering
\includegraphics[width=\linewidth]{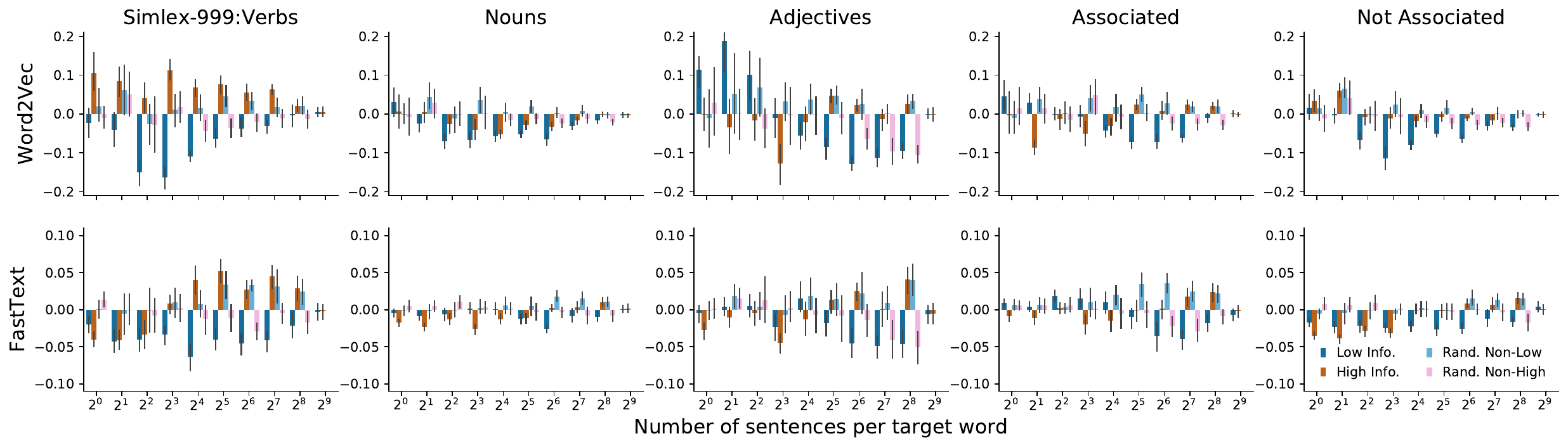}

\caption{
Performance differences between curricula from the random curriculum in different POS and association pairs of SimLex-999 target words. 
Notice scales are different between Word2Vec and FastText models. 
}
    \label{fig:curriculum_result_simlex_details_diff}
\end{figure*}

Figure \ref{fig:curriculum_result_simlex_details_diff} shows the comparison between curricula from the random curriculum by different POS and association pairs of SimLex-999 target words. 
We can observe similar patterns from other similarity tasks' results, as the low informative models perform worse than other curriculum and the non-low informative sentence models perform better than other cases.

\paragraph{By Associated Pairs}
\label{sec:exp3__results_assoc}

\begin{figure}[t]
\centering
% \hspace*{-0.61in}
\includegraphics[width=0.67\linewidth]{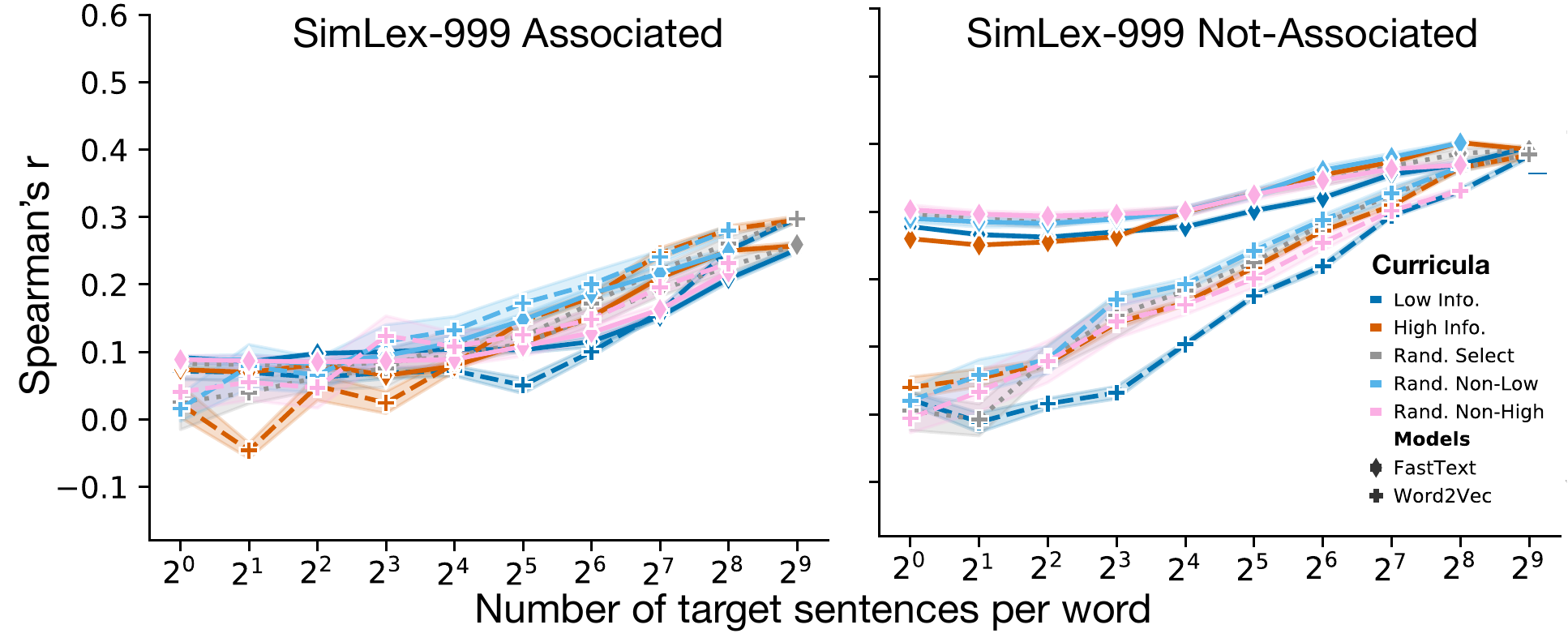}
\caption{
Batch learning results on Word2Vec\textsuperscript{\sdiam} and FastText\textsuperscript{$\bullet$} by association SimLex-999 target word pairs. 
The non-low informative sentence models (\texttt{Rand. Non-Low}) performed better than other models. Unlike other results, FastText models did not show an early learning advantage with associated target words.
}
    \label{fig:curriculum_result_simlex_assc}
\end{figure}

% We also compared the results by the associated and non-associated pairs in SimLex-999 task. 
% We observed that curricula developed based on our contextual informativeness scores are able to consistently distinguish between less vs. more helpful training sentences for Word2Vec and FastText models (Figure \ref{fig:curriculum_result_simlex_assc}). 
SimLex-999 also contains indicators for associated and non-associated pairs. For example, associated pairs include words like \textit{car--petrol}, or \textit{old--new}. 
Although the associated words are related to each other, it does not mean they are highly similar words. Thus, accurately measuring the similarity scores of the associated word pairs is considered as a harder task for word embedding models, since the contexts can be similar for each word in associated pairs~\citep{hill2015simlex}.

With the non-associated target word pairs from SimLex-999 task, the non-low informative random sentence models (\texttt{Rand. Non-Low}) performed in comparable or better levels than the random sentence models (\texttt{Rand. Select}) (Figure~\ref{fig:curriculum_result_simlex_assc}). The high-informative models (\texttt{High Info.}) showed consistently better performance than the low-informative models (\texttt{Low Info.}). 

For the associated pairs, Word2Vec models trained with the non-low informative random sentences (\texttt{Rand. Non-Low}) performed best. Surprisingly, FastText models did not show early learning advantages, showing that the associated pair task is a hard task for both Word2Vec and FastText word embedding models. 

% Using enough high-informative sentences (e.g., $> 2^5$ sentences per target word) provided better performance than the low informative models and the random sentence models. Similar to the adjective pair results, the low-informative sentence models showed a moderate level of performance with a smaller number of sentences per target word, but the performance stagnated until the number of training sentences reached a higher number (e.g., $2^6$ sentences per target word). On the other hand, the performance of non-low informative sentence models improved continuously. 

% The results for analyzing relatedness and similarity scores separately~\citep{agirre2009study} did not indicate different patterns from the overall results. 
% These results indicate that our model for predicting contextual informativeness scores works well in distinguishing less vs. more helpful training sentences for the Word2Vec model. 

%%%%%%%%%%%%%%%%%%%%%%%%%%%%%%%%%%%%%%%%%%%%%%%%%%%%%%%%%%%%%%%%%%%%%%%%%%%%%%%%%%%%%%%%%%%%%%%%%%%%%%%%%%%%%%%%%%%

\section{Experiment 4: Curriculum Learning for Few-Shot Learning Model}
\label{sec:exp4}
In Experiment 3, we examined the curriculum learning effect, based on contextual informativeness, for word embedding models in a batch learning setting. 
In Experiment 4, we further investigated the effectiveness of curriculum design with a few-shot learning model, which the quality of dataset can be more critical.
We used \textit{Nonce2Vec} model~\citep{herbelot2017high} to test this with model intrinsic metrics and comparison with annotated dataset.

\subsection{Models, Dataset, and Task}
\label{sec:exp4__model_dataset_task}
% - heuristics for curriculum building

\subsubsection{Few-Shot Learning Model}
For this experiment, we used Nonce2Vec, where the word embedding model learns new target words with a very small number of sentences. 
% , which is a variant of Word2Vec that specializes in learning with a smaller training corpus for nonce words. 
The model is a variant of Word2Vec, and employs a higher initial learning rate and customized decay rate, to provide a riskier but more effective learning strategy for unseen words with a small number of training sentences. 
We followed a similar experimental setup for training Nonce2Vec as in \cite{herbelot2017high}. 
The few-shot learning setting also first trained a background model and then tested the effect of adding new sentences by updating the model.
For a background model, we trained a regular Word2Vec model with the same parameters from Experiment 3.
% This is a different setting from the batch learning models. 

Unlike the word embedding models in Experiment 3, the background model of Nonce2Vec was trained with \textit{both} target and non-target sentences.  
% Nonce2Vec model takes the nonce word (i.e., target word) as an input. 
The model simulates first-time exposure to the target word by changing the label of the existing target word's vector from the background model, and adds a newly initialized word vector for the target word. 
For example, if the target word is \textit{insulin}, Nonce2Vec copies the target word's vector from the background model with a different label, like \textit{insulin\_gold}. Then the model randomly initializes the original target word's vector, and learns the new vector representation for the target word with a small number of target sentences~\citep{herbelot2017high}.
As further described in Section \ref{sec:exp4__evaluation_task}, we could compare the original word vector as the gold-standard with a new word vector learned from few-shot examples.

\subsubsection{Building Curricula}
\label{sec:exp4__dataset_curriculum}
% However, the experiment setting is little different from Experiment 3 (Section \ref{sec:exp3}). 

For this experiment, the background corpus contained non-target sentences \textit{and} $\sim$60\% of randomly selected target sentences per each target word. Training curricula were derived from the other 40\% of the target sentences and used to update the nonce word vectors. 

Following the previous study~\citep{herbelot2017high}, we used 2, 4, or 6 target sentences per target word to train the model. 
% However, using only a small number of sentences may increase the risk of developing curriculum based on incorrect informative score prediction.
For a more robust comparison, we selected each curriculum stochastically. For example, we first created a sampling pool of 50 sentences (about 10\% of sentences per target word) for each curriculum type (with overlaps).
For non-low/non-high informative curricula, we first filtered out 256 low/high informative sentences and then sampled 50 sentences from the rest.
With some overlapping sentences, the overall number of target sentence pool was about 40\% of total target sentences per target word. 
From each pool, we then randomly sampled 2, 4, or 6 target sentences per target word to develop the curriculum for each iteration. 

\subsubsection{Evaluation Tasks}
\label{sec:exp4__evaluation_task}
% - model inherent evaluations (e.g., median rank, perplexity?)
% - Semantic similarity datasets
% We also adopted evaluation methods from~\cite{herbelot2017high}.
For evaluation, we used the SimLex-999 task, as the task includes word pairs with various lexical attributes (e.g., association or part of speech).  % for the target words and annotated semantic similarity scores, 
First, to evaluate the quality of nonce word vectors, we measured the median rank of newly-learned nonce word vectors and the gold-standard word vectors included in the background model. 
Ideally, if the nonce word learning process of Nonce2Vec is perfect, the embedding vector for the nonce word and the gold-standard word from the background model (e.g., \textit{insulin} vs. \textit{insulin\_gold}) should be identical ($rank=1$). Similarly, lower median rank scores for the target words (e.g., SimLex-999) would indicate better embedding quality derived from a curriculum. 

Second, we also compared Spearman's rank correlation scores of each semantic similarity task. 
For Nonce2Vec models, we took the average of two cosine similarity scores for a word pair, as we conducted the nonce learning for each target word separately. For example, for the word pair \textit{old--new}, we first conducted nonce word learning for the word \textit{old} and calculated the cosine similarity score with the word \textit{new} from the background model. Then we conducted the same process vice versa. The average of these two cosine similarity scores was calculated for the word pair.

\subsection{Results}
\label{sec:exp4__results}
\subsubsection{Nonce Word Vector Ranking}

\begin{figure}[t]
\centering
\includegraphics[width=0.67\linewidth]{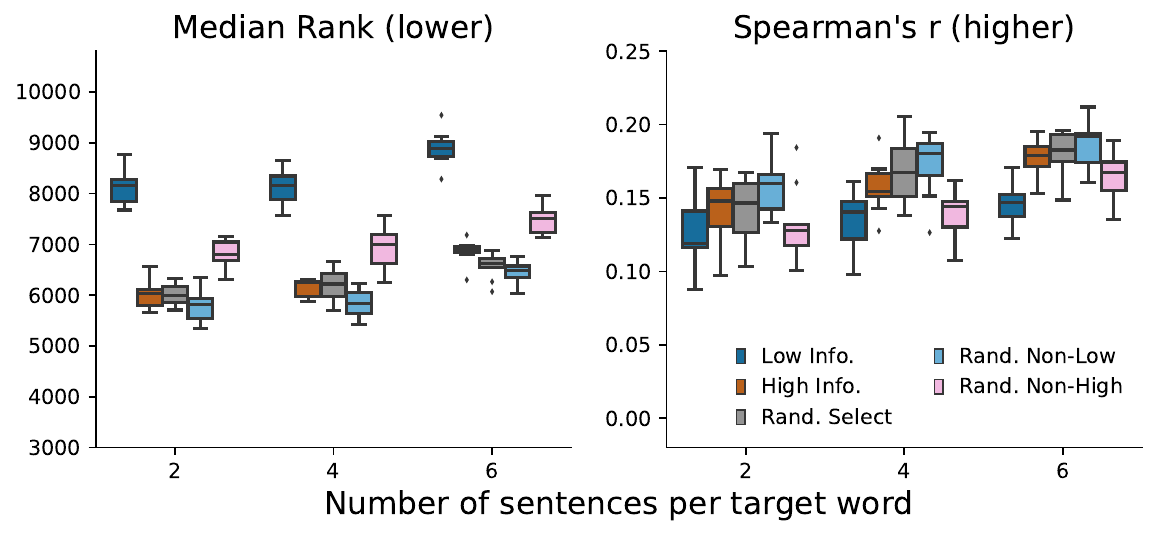}

\caption{
Few-shot learning results (Nonce2Vec) on SimLex-999 dataset.
In terms of embedding quality, measured in median rank (lower is better), the low-informative sentence (\texttt{Low Info.}) and non-low informative sentence curricula (\texttt{Rand. Non-Low}) provided the worst and best embedding quality respectively.
% , compared to the gold-standard word vectors. 
Spearman's \textit{r} score (higher is better) tended to increase as the number of sentences per target word increased, while there were no significant differences observed between curricula. 
}
\label{fig:curriculum_result_simlex_n2v}
\end{figure}

% Figure \ref{fig:curriculum_result_simlex_n2v} shows how Nonce2Vec models ($epoch=5$, $learning rate=0.5$) performed with different curricula. 
Figure \ref{fig:curriculum_result_simlex_n2v} shows that randomly selecting sentences that are not low-informative (\texttt{Rand. Non-Low}) produced consistently better performance at the few-shot word learning scenario. We could observe similar patterns from the batch learning results. 
% We compared the quality of updated nonce word vectors by comparing the similarity ranking with the gold-standard word vector. 
% Our results show that randomly selecting sentences from the non-low informative curriculum produced consistently better performance at learning word meaning in a few-shot learning scenario.
The median rank scores indicate that the low-informative sentences (\texttt{Low Info.}) performed significantly worse than other curricula.
% , while, the non-low informative random sentences (\texttt{Rand. Non-Low}) tended to perform better than others. 
The high-informative sentence models (\texttt{High Info.}) showed significantly better results than the low-informative sentence models, but not much different from the randomly selected sentence models (\texttt{Rand. Select}). 

These results show that our contextual informativeness model can effectively distinguish less-helpful sentences for the few-shot learning tasks. 
Moreover, based on the random non-low results, we see that excluding the least informative examples, together with using diverse levels of contextual informativeness stimuli (medium and high), can improve word embedding model performance.

\subsubsection{Semantic Similarity}
% Overall, the lower learning rate (0.5) models showed significantly better correlation scores. 
Also, we analyzed the rank correlation between the Nonce2Vec word vectors and human-annotated scores from SimLex-999. 
The results from Spearman's \textit{r} score showed similar patterns from the median rank results. However, we did not observe significant differences between curricula. 
The low-informative sentence models (\texttt{Low Info.}) performed marginally worse than other curricula. 
The non-low informative models (\texttt{Rand. Non-Low}) and randomly selected sentence models (\texttt{Rand. Select}) performed similarly. 

% We could also observe that using more sentences per target word marginally improve the overall performance of SimLex-999 correlation task. 
% Correlation scores were also tend to improve with a larger number of sentences per target word.
% However, we could not observe significant differences between sampled high and low informative sentence curricula.
% In this analysis, we could not observe significant differences between sampled high and low informative sentences. 

%%%%%%%%%%%%%%%%%%%%%%%%%%%%%%%%%%%%%%%%%%%%%%%%%%%%%%%%%%%%%%%%%%%%%%%%%%%%%%%%%%%%%%%%%%%%%%%%%%%%%%%%%%%%%%%%%%%

\section{Discussion and Limitations}
% - where and why model did/not worked well
%     - contextual informative prediction
%     - curriculum learning results
% - potential applications

We view our study as bridging recent deep learning advances in semantic representation with educational applications. 
Predicting, characterizing, and ultimately, optimizing the quality of a learner's initial encounters with content has many potential applications for both human and machine learners. 
Our results showed that 1) our attention-based deep learning model can be effective for predicting the instructional aspects of contextual informativeness, 2) that our model can also provide interpretable results on how context words are used to infer the meaning of the target word, and 3) identifying low-informative sentences and filtering those out from the training set can significantly improve the quality of word embedding models, as measured across various sentence sizes and similarity tasks. 
For human learners, contextual informativeness models could be applied to diverse sources of classroom-generated text, such as video transcripts or class notes, to find the most supportive lecture contexts that help learn or review specific terminology. Search engines could emphasize the most contextually informative educational Web content examples for a given term or concept. Using custom pre-trained models, like BioBERT~\citep{lee2019biobert} for biomedical terms and content, would enable more domain-specific applications.

On the algorithmic side, accurate prediction and characterization of contextual informativeness would be highly valuable in NLP applications, including finding more supportive sentences for automatic summarization~\citep{kaageback2014extractive,mihalcea2004textrank}, or automatic curriculum design for few-shot learning NLP tasks~\citep{herbelot2017high}, where the quality of the training examples is critical. 
Also, our new single-sentence dataset could be a valuable resource for evaluating few-shot learning capabilities, such as training sophisticated language modeling systems~\citep{gp3_2020}, relevant language representations~\citep{wang2017distributional}, or generative language patterns~\citep{lake2015human}.

We also discuss here a few limitations of this study that may inspire further research. 
A more complete model of contextual informativeness would include an individual component capturing a specific user's background knowledge of the target concept, although such models can be challenging to train and evaluate.
For example, individual differences in word knowledge~\citep{borovsky2010learning} or language proficiency~\citep{elgort2018contextual} may result in different levels of comprehension or faster processing of orthographic information. 
While collecting BWS annotations for the single-sentence contexts, we tried to minimize these confounds by collecting multiple responses per sentence and limiting the geographic location of annotators to native English speaking countries. 
However, further comparison between different learner profiles, such as L1 vs. L2 learners, could benefit developing more personalized and/or group-oriented prediction. 

Our attention-based model showed significant improvements over the baseline models, but still has some room to improve. 
For example, in Section \ref{sec:exp1__res_single}, the BERT baseline model did not perform significantly better than other simpler baseline models that used sentence length or co-occurrence information. 
We also observed that prediction results may be somewhat less accurate for lower- and higher-end informative scores (Appendix~\ref{app:residual_analysis}). 
% For both single- and multi-sentence contexts, our model tended to over-predict the informative scores of low-informative contexts, and under-predict the high-informative contexts' scores. 
Although we did not identify any systematic characteristics of these less accurate prediction cases, the results might indicate the pre-trained NLP model's limited vocabulary size or difficulty in processing grammatically complex or incorrect sentences. 
Example-based analysis, such as word-level permutation, to investigate which context words most impact prediction results~\citep{kaushik2019learning}, can help to systematically identify these difficult cases. 
Applying more sophisticated methods for processing attention weights from multiple layers~\citep{abnar2020quantifying} may provide more accurate results.
\cite{jiang2019how} points out that the performance of pre-trained models on predicting cloze responses can be affected by specific contextual terms in a sentence. Our future work could evaluate our prediction model with diverse sentences for a more thorough comparison. 
More thorough hyperparameter optimization process would improve the models' performance too.

% Also, we relied on a single attention layer to learn relationships between the context words and the target word. 
% The limited lexical complexity of sentences in the EVALution dataset~\citep{santus2015evalution} may not provide a full view of attention weight quality. Although BERT-based models performed better than ELMo-based models for multiple prediction tasks, they gave lower quality attention distributions on capturing the pair words (Section \ref{sec:attention_eval}).  

In curriculum learning experiments, Word2Vec or FastText model performance reached a plateau relatively quickly (e.g., $\geq 2^8$ sentences per target word (Section \ref{sec:exp3__results}). Bigger models, such as transformer- or LSTM-based models, might have more headroom to test the curriculum effect with a larger number of target sentences, and with different downstream tasks such as question and answering or summarization. 

Further investigation is needed to understand the details of curriculum effects. Our results in Section~\ref{sec:exp3__results} showed that different linguistic properties of target words like POS or association can introduce different training results, possibly due to different grammatical relationships between context words and the target word. 
Some context words that often co-occur with the target word may be more informative~\citep{kabbach2019towards}. 
High-informative sentences may contain fewer redundant or frequent context words. 
More detailed analysis on where the BERT-based contextual informativeness model fails in predicting contextual informativeness scores could improve the derived curriculum. 
Identifying what accounts for the curriculum effects could provide more specific guidelines for human instructors and algorithm designers. 

We can try more sophisticated curriculum development strategies. 
As previous studies in vocabulary acquisition~\citep{frishkoff2016dynamic} and machine learning~\citep{sachan2016easy} suggest, developing more effective curricula can be non-trivial. 
Different levels of informative contexts have different roles in learning~\citep{van2018contextual}. As we observed from Sections~\ref{sec:exp3__results} and~\ref{sec:exp4__results}, the non-low informative sentence models, which filtered out the less-informative target sentences but still kept diverse levels of informativeness scores, performed better than the high-informative sentence models in almost every case. 
Developing more sophisticated learning strategies, such as using high-informative sentences for the initial higher loss state and low-informative sentences in a more mature model state, could be an interesting curriculum learning problem and opportunity to compare with human students learning in a scaffolded condition. 

Lastly, our efforts focused on developing English sentences because our students were English-language learners. 
Further studies using non-English datasets, accounting for more inflected languages or complex grammatical rules, or semantic biases associated with different cultures ~\citep{osgood1957measurement,bolukbasi2016man} 
would be a valuable complement to the science of reading literature.

\section{Conclusion}
Both humans and machines rely on context for acquiring a word's meaning---yet not all contexts are equally informative for learning. 
In Experiment 1, we introduced and evaluated a novel attention-based deep learning model that effectively captured a general conception of contextual informativeness, as demonstrated by its successful prediction performance and its ability to generalize across significantly different datasets. Our model showed that deep neural representations learned automatically could compete well with a complex, feature-engineered model, and that combining the best features of both methods led to superior, state-of-the-art accuracy overall.
In Experiment 2, we demonstrated that learned attention mechanisms can provide interpretable explanations that match human intuition as to which specific context words help human readers infer the meanings of new words.

In Experiments 3 and 4, we showed that curricula selected by contextual informativeness scores can help improve the representations learned by various types of word embedding models. 
In a series of experiments across batch learning and few-shot learning, we tested the effectiveness of rating contexts by informativeness to prioritize those contexts likely to aid to learning word meaning.
Our results show that sentences predicted as low-informative by the model are indeed generally less effective for training word embedding models. Further, in most cases filtering out low-informative sentences from the training set substantially improves word representations for downstream tasks. In the future, we will further investigate how to build optimized curricula for word learning, and identify factors related to the curriculum effect based on contextual informativeness.

\backmatter

% \bmhead{Supplementary information}
% `Not applicable'
% If your article has accompanying supplementary file/s please state so here. 

% Authors reporting data from electrophoretic gels and blots should supply the full unprocessed scans for key as part of their Supplementary information. This may be requested by the editorial team/s if it is missing.

% Please refer to Journal-level guidance for any specific requirements.

% \bmhead{Acknowledgments}
% `Not applicable'
% Acknowledgments are not compulsory. Where included they should be brief. Grant or contribution numbers may be acknowledged.

% Please refer to Journal-level guidance for any specific requirements.

\newpage
\begin{appendices}

\section{Generating and Annotating Single-Sentence Contexts} 
% dscovar-tr-2015-008
\label{app:sent_gen}
For the single-sentence context dataset, Our researchers manually generated sentences in different contextual informativeness levels with respect to the target word. 
We provided general instructions (Table \ref{tab:sent_gen_inst}) for creating low, medium, and high informative cloze sentences. 
We also included example sentences and descriptions \ref{tab:sent_gen_good}) in different contextual informativeness levels and target words' part-of-speech to help researchers to create cloze sentences correctly. 
We additionally provided example phrases (Table \ref{tab:sent_gen_phraselow}) and poor example sentences (Table \ref{tab:sent_gen_poor}) for low contextual informative cases, to further control the quality of cloze sentences.

\begin{table*}[!htbp]
\centering
\begin{tabular}{p{0.7cm} p{10.2cm}} \hline
Lv.  & Instructions \\ \hline
High& 1. Use simple sentence structures. Note that target grade level for these contexts is 4th grade. All sentences should be between 9 and 13 words long (mean ~11).\\
    & 2. Use only easy, familiar words (except for target of course). Note that the average grade level for our contexts in the previous study was 4th grade.\\ % KiWL1
    & 3. Target word should be placed towards the end of the sentence whevever possible. You may find that this requirement completes with \#1. Do your best.\\
    & 4. Refer to list of synonyms, near synonyms, and cohorts.\\
    & 5. Remember to avoid difficult (Tier 2) words. Look for simple synonyms and related words using LSA and/or a good thesaurus.\\
    & 6. Each sentence should work with BOTH the very rare words and their Tier 2 synonyms. Please take the time to try to understand correct usage. Use the following website to look up actual usage of words if you're not sure:  http://www.onelook.com (Onelook.com links to two sites that are particularly helpful: Vocabulary.com and https://www.wordnik.com, which provide a good range of example sentences, as well as definitions and explanations of correct usage).\\
Med & (in addition to 1-6 above) 7. It is not easy to know a priori whether a sentence will turn out (based on cloze data) to be medium or high constraint. In addition, this classification depends on the metric (e.g., whether we're looking at lexical/cloze data or at a derived measure that captures semantic constraint).Nonetheless, I've tried to assemble what I think are good a priori examples of Med vs. High C. Don't sweat the difference too much. We'll need cloze data to determine which way they fall.\\
Low & 1. Use simple sentence structures. Note that target grade level for these contexts is 4th grade. All sentences should be between 9 and 13 words long (mean ~11).\\
    & 2. Use only easy, familiar words (except for target of course). Note that the average grade level for our contexts in the previous study was 4th grade.\\ % KiWL1
    & 3. Target word should be placed towards the end of the sentence whevever possible. You may find that this requirement completes with \#1. Do your best.\\
    & 4. For low-constraint sentences, avoid any content words (adjectives, nouns, or verbs) that could prime specific concepts. \\ \hline
\end{tabular}
\caption{The instructions for generating cloze sentences in different contextual informativeness levels.}
\label{tab:sent_gen_inst}
\end{table*}

\begin{table*}[!htbp]
\centering
\begin{tabular}{p{0.7cm} p{0.7cm} p{9.0cm}} \hline
Lv.     & POS   & Example sentence / Descriptions \\ \hline
High    & Noun  & \textit{We covered our ears to block the loud $\rule{7mm}{0.15mm}$ from the \textbf{crowd}.} \\
        &       & Bolded words all prime the concept `noise'. `Ears' and `loud' are super constraining. Try out Hi-constraining sentences on a few friends, colleagues to ensure that they don't come up with other concepts you didn't think of when you created sentence. The cloze probability for `noise' was 70\% \\
        & Adj   & \textit{Wendy \textbf{used to be fat, but} after her \textbf{illness} she looked $\rule{7mm}{0.15mm}$.} \\ 
        &       & `but' indicates that missing word is opposite of `fat' (thin, skinny, gaunt). Note that `illness' suggests negative rather than positive characteristic (so `skinny' or `gaunt' were both common responses). The joint cloze probability for `thin' + `skinny' was $>$70\%. \\
        & Verb  & \textit{The burglar was caught by police while trying to $\rule{7mm}{0.15mm}$ the jewelry.} \\
        &       & `burglar', `police' suggest criminal activity; `try to' suggest activity was thwarted, so `buy' for example would be pragmatically odd. The cloze probability for "steal" was around 70\%. \\
Med     & Noun  & \textit{I enjoyed my flight to Paris except for all the $\rule{7mm}{0.15mm}$.} \\
        &       & `enjoy $\rule{7mm}{0.15mm}$' or `except $\rule{7mm}{0.15mm}$' suggests that $\rule{7mm}{0.15mm}$ refers to something undesirable/unpleasant. On a flight there are usually lots of people, incl. screaming babies \& small children $\rightarrow$ `noise'. However, there is also `turbulence' and `delays', two other concepts that were also provided by several respondents on the cloze task. The cloze probability for `noise' was 13\% (cloze for `turbulence' was $\sim$25\%) \\
        & Adj   & \textit{The doctor warned the woman she was too $\rule{7mm}{0.15mm}$ from a poor diet.} \\
        &       & Could be `thin' or `skinny,' but also `fat' or `sick.' The joint cloze probability for `thin' + `skinny' was 29\%. \\ 
        & Verb  & \textit{Sam said he would have a million dollars if he $\rule{7mm}{0.15mm}$ it.} \\
        &       & People said `invest', `earn', and `save' as often as they said `steal.' The cloze probability for `steal' was around 30\%. \\
Low     & Noun  & \textit{The group \textbf{did not choose that one} because of all the $\rule{7mm}{0.15mm}$.} \\
        &       & `did not choose' is not very constraining since we don't know what `that one' refers to. \\
        & Adj   & \textit{I was \textbf{surprised} to find that \textbf{it} was $\rule{7mm}{0.15mm}$.} \\
        &       & `surprise' doesn't suggest anything in particular about the characteristics of the noun (it); `it' can refer to almost anything. \\
        & Verb  & \textit{We were \textbf{interested} to learn that Sally had \textbf{decided not to} $\rule{7mm}{0.15mm}$ .} \\
        &       & `interested to learn' is could refer to almost anything; same with `decided not to $\rule{7mm}{0.15mm}$.' \\ \hline
\end{tabular}
\caption{The list of good cloze sentence examples with descriptions used for generating cloze sentences with different semantic constraint levels.}
\label{tab:sent_gen_good}
\end{table*}

\begin{table*}[!htbp]
\centering
\begin{tabular}{r p{10cm}} \hline
POS     & Example Phrases \\ \hline
Noun    & $\rule{7mm}{0.15mm}$ appeared/disappeared, turn into $\rule{7mm}{0.15mm}$, $\rule{7mm}{0.15mm}$ came into sight, think about/imagine $\rule{7mm}{0.15mm}$, found/lost/discovered $\rule{7mm}{0.15mm}$, remember/recall $\rule{7mm}{0.15mm}$, buy/lend $\rule{7mm}{0.15mm}$, write about $\rule{7mm}{0.15mm}$, have $\rule{7mm}{0.15mm}$, forget about $\rule{7mm}{0.15mm}$ \\
Adj     & is/seems/looks/appears $\rule{7mm}{0.15mm}$, decide/judge whether it is $\rule{7mm}{0.15mm}$, become/turn $\rule{7mm}{0.15mm}$, think/believe/know it is $\rule{7mm}{0.15mm}$, say/argue it is $\rule{7mm}{0.15mm}$ \\
Verb    & decide to $\rule{7mm}{0.15mm}$, hard/impossible to $\rule{7mm}{0.15mm}$, used to $\rule{7mm}{0.15mm}$, $\rule{7mm}{0.15mm}$ more/less often (in the future), agree to $\rule{7mm}{0.15mm}$, see/watch someone $\rule{7mm}{0.15mm}$, try to $\rule{7mm}{0.15mm}$, have/need to $\rule{7mm}{0.15mm}$, imagine what it's like to $\rule{7mm}{0.15mm}$, learn (how) to $\rule{7mm}{0.15mm}$, remember (how) to $\rule{7mm}{0.15mm}$ \\ \hline
\end{tabular}
\caption{Example phrases provided for generating low constraining sentences.}
\label{tab:sent_gen_phraselow}
\end{table*}

\begin{table*}[!htbp]
\centering
\begin{tabular}{r p{10cm}} \hline
POS     & Example sentence / Descriptions \\ \hline
Noun    & \textit{Morgan did not like Bob because she thought he was a(n) $\rule{7mm}{0.15mm}$.} \\
        & `did not like' suggests that the target word refers to something undesirable/displeasing; linking target word to `Bob' may bias reader to think of `male' traits, occupations. \\
Adj     & \textit{I was disappointed to find that the boat was $\rule{7mm}{0.15mm}$.} \\
        & `disappointed' suggests that the target word refers to something undesirable; `the boat' is much too constraining (the target word can only refer to properties of boats). \\
Verb    & \textit{We were thrilled to learn that Sally had decided not to $\rule{7mm}{0.15mm}$ .}  \\
        & `thrilled to learn' has positive connotations (so it increases the constraint) \\ \hline
\end{tabular}
\caption{The poor examples for the low informative sentences with descriptions.}
\label{tab:sent_gen_poor}
\end{table*}

\subsection{Iterative Refinement of Contexts}
Most of our target words were lower-frequency, \textit{Tier 2} words that are critical in writing but are rarely encountered in everyday speech. It is therefore not surprising that researchers (even those with excellent vocabulary) sometimes generated contexts that misuse target words. This might happen because researchers relied on the synonym and cohort words to come up with new examples, rather than retrieving them from high-quality published sources. In any case, new human-generated contexts must be vetted for correct usage.  
% Previously, we have asked at least 2-3 people (including Dr. Frishkoff) to vet each context for correct usage. This is time-consuming, but necessary.
% Ideally, we would start with examples from high-quality, published sources and iteratively refine them to meet our readability, length, and other criteria. We are beginning to explore this process in collaboration with Maxine Eskenazi at CMU. This work will be documented in a forthcoming TR.

% \subsection{Validation of Context Constraint}
% The precise methods for validating level of contextual informativeness are documented in a forth-coming TR. We start by collecting "cloze" responses for each context (removing target word and asking native English speakers to fill in the blank). We typically collect around 30-50 participant responses for each context. We then compute two kinds of informativeness ratings for each context. Lexical constraint is estimated by determining the number of distinct cloze completions (weighted by frequency of each response). Semantic constraint is estimated in the same way, but the responses are first clustered based on measures of semantic simlarity. The clusters can be defined narrowly, such that each cluster includes words that are close synonyms, or more broadly, so that clusters represent hyperonymic categories (e.g., "food," "happy emotions") that include words that are semantically related, but not strict synonyms. 

\subsection{Annotating the Single-Sentence Context Dataset} \label{app:score_cloze}
All crowdsourcing tasks were conducted on the Figure Eight\footnote{\url{https://www.figure-eight.com/}} platform.  Two types of crowdsourcing annotation were performed.
Workers were shown 10 tuples at a time (per page) of annotation and paid \$0.25 per page. To ensure the quality of annotations, one control question that had a known judgment (e.g., being the most informative sentence of a tuple) was randomly inserted per page. Workers were required to maintain at least 80\% accuracy on these control questions during annotation to continue annotating.

We sampled sentences using the following criteria based on~\cite{kiritchenko2016capturing}.
\begin{itemize}
    \item No two $k$-tuples have the same $k$ terms.
    \item No two terms within a $k$-tuple are identical.
    \item Each term in the term list appears approximately in the same number of $k$-tuples.
    \item Each pair of terms appears approximately in the same number of $k$-tuples.
\end{itemize}

From the entire questionnaire, each sentence appeared in 8 different tuple sets  ($2 \times n$, 4-tuple sets were included in the entire questionnaire set by following the previous study~\citep{kiritchenko2016capturing}). 
% In total, each sentence appeared as an option of the tuple set 24 times. 

\begin{figure*}[!htbp]
% \hspace*{-35mm}
\centering
\includegraphics[height=\textheight]{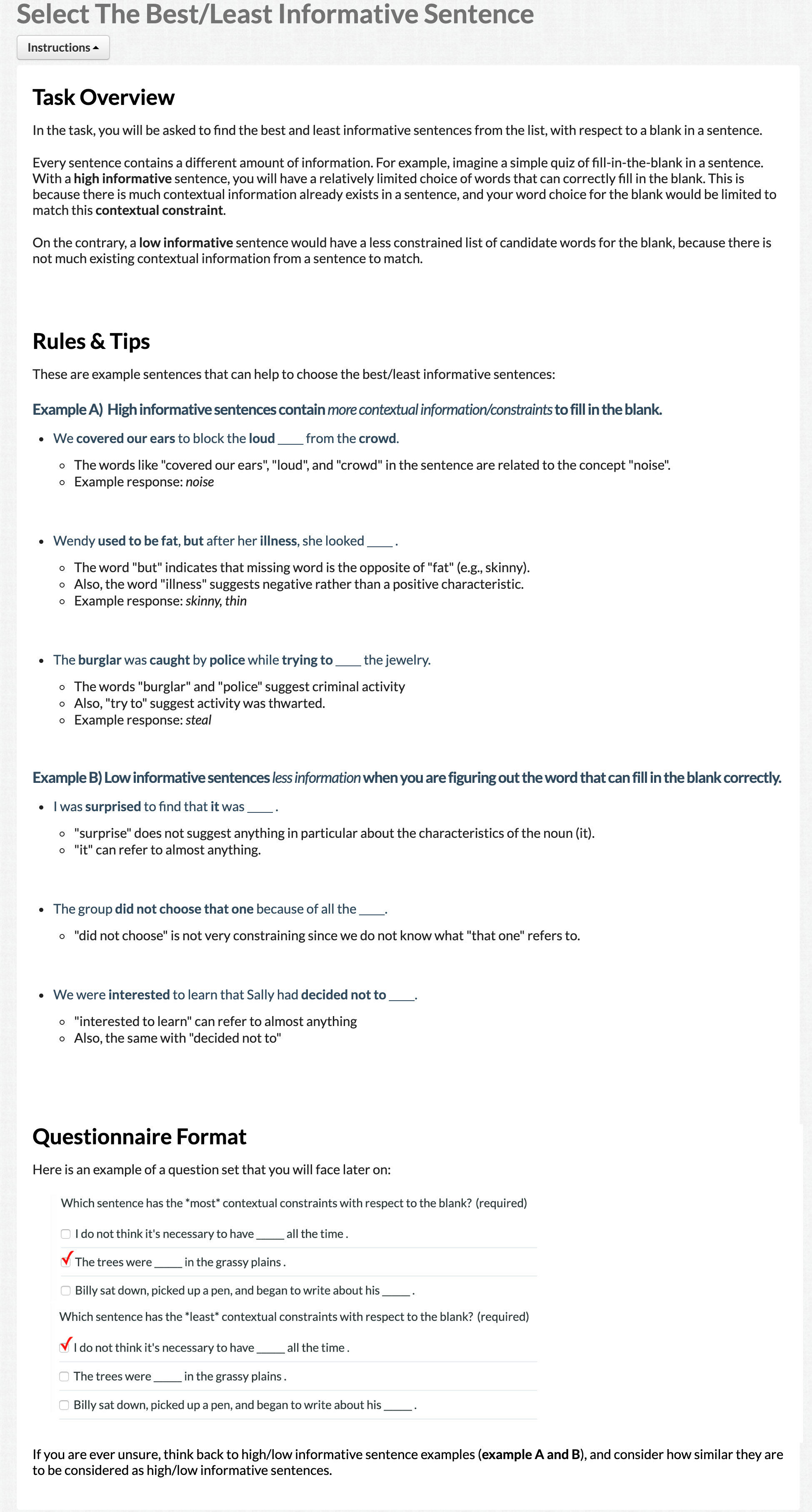}
\caption{Instructions used for collecting cloze sentence annotations.}
\label{fig:crowdsourcing_t2_scoring}
\end{figure*}

Figure \ref{fig:crowdsourcing_t2_scoring} includes the instruction page that we presented to crowdworkers. It includes the definition of high or low contextual informativeness with examples. 
Inter-rater agreement rates for the best and worst sentence picks for each tuple were 0.376 and 0.424 with Krippendorff's $\alpha$. Although these scores scores are in moderate level, we view this agreement is still acceptable given the replicability score across random splits is similar from the previous studies~\citep{kiritchenko2016capturing,kiritchenko2017best}. 

\section{Attention-Based Contextual Informativeness Prediction Model}
\subsection{Hyperparameters}
\label{app:hyperparams}
During the training process of ELMo- and BERT-based models, we fine-tuned the pre-trained models.
Because of the differences in the number of trainable parameters of each pre-trained model, we used different learning rates for each ELMo-based ($1\mathrm{e}{-3}$) and BERT-based ($3\mathrm{e}{-5}$) model. Other hyper-parameters remained constant across models (batch size: 16, iteration: 5 (for the single-sentence) or 3 (for the multi-sentence context dataset). 
These hyperparameters were selected based on the preliminary study. 
The dimensions for the attention block layers used same dimensions with the pre-trained embeddings (ELMo: 1024, BERT: 768). 
 The dimension of the linear layer in the regression block was 256. 

For the baseline model using co-occurrence information, the ridge regression model was trained with \texttt{scikit-learn}'s default alpha value. Co-occurrence matrix was built for words that appeared more than five times in the training data. 

The replicated random forest model from~\citep{kapelner2018predicting} followed the original paper's setting. We used features and settings provided by the authors, setting the number of estimators as 500 and bootstrapping sample size as 10000, and including 600+ lexical features that included n-gram frequencies from Google API, and Coh-Metrix~\citep{mcnamara2014automated}, sentiment~\citep{crossley2017sentiment}, psycholinguistic~\citep{crossley2016tool}, and other lexical sophistication features~\citep{crossley2016tool,kolb2008disco}.

\subsection{Computing Resource for Training}
For this study, we used a single NVIDIA 2080 TI GPU with Intel i7 CPU. For training the model with the single-sentence context dataset, it took about 1 minute per fold (90\% of the data). For the multi-sentence context dataset, it took approximately 30 minutes per fold. 

We used pre-trained versions of ELMo~\citep{peters2018deep}
and BERT~\citep{devlin2019bert} from TensorFlow Hub~\url{https://tfhub.dev/}. 
The ELMo-based model with attention block had about 426k trainable parameters, while the BERT-based counterpart had about 7.3M trainable parameters. 

%%%%%%%%%%%%%%%%%%%%%%%%%%%%%%%%%%%%%%%%%%%%%%%%%%%%%%%%%%%%%%%%%%%%%%%%%%%%%%%%%%%%%%%%%%%%%%%%%%%%%%%%%%%%%%%%%%%
\section{Results: Experiments 1 and 2}
The following sections include additional analysis results that we did not included in Sections~\ref{sec:exp1__res_single} and~\ref{sec:exp1__res_multi}.

\subsection{Residual Analysis}
\label{app:residual_analysis}
\begin{figure}[th]
\centering
\includegraphics[width=0.6\linewidth]{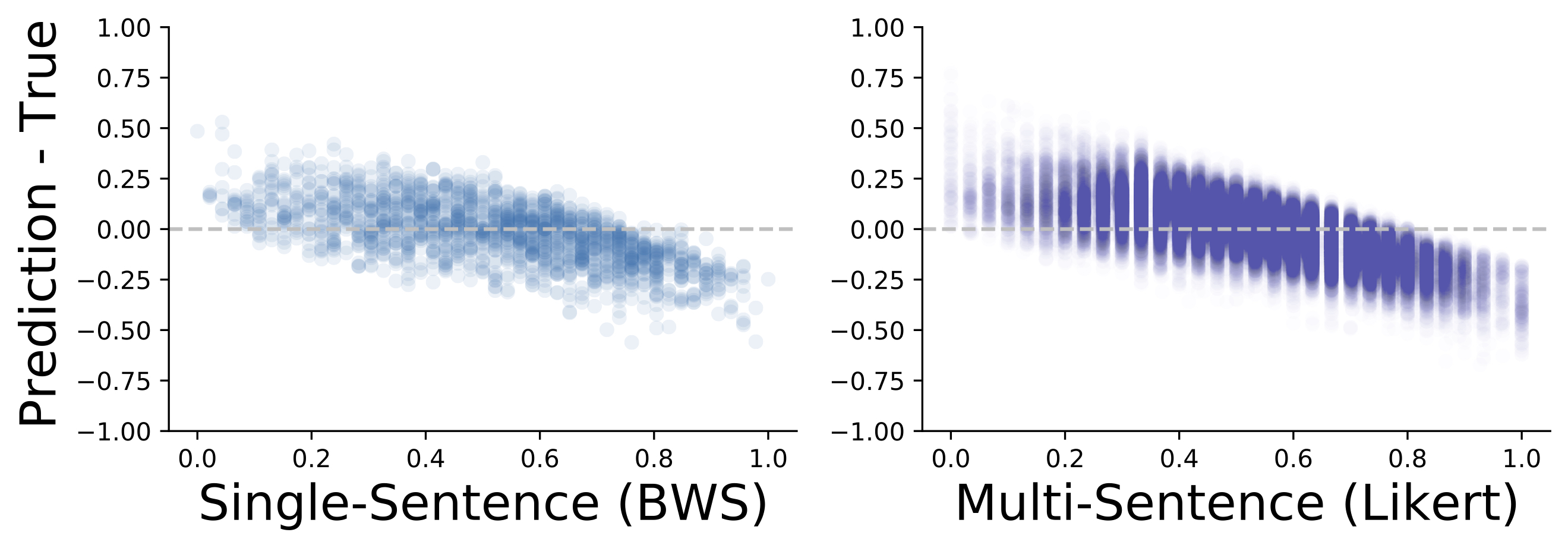}

\caption{
Residual of predicted scores compared to the true score of single- and multi-sentence contexts.
For both datasets, our model over-predicted the informative scores for low informative sentences, and under-predicted the high informative sentences. 
}
    \label{fig:residual_analysis}
\end{figure}

Figure \ref{fig:residual_analysis} shows that our model over-predicted the informative scores for low informative sentences, and under-predicted the high informative sentences. 

%%%%%%%%%%%%%%%%%%%%%%%%%%%%%%%%%%%%%%%%%%%%%%%%%%%%%%%%%%%%%%%%%%%%%%%%%%%%%%%%%%%%%%%%%%%%%%%%%%%%%%%%%%%%%%%%%%%

\subsection{Single-Sentence Contexts}
\label{app:exp1_results_single}
Table \ref{tab:pred_dscovar} shows more details of our models' performance on both semantic density and BWS informative scores of the single-sentence dataset. 

% \begin{sidewaystable}
% \sidewaystablefn%
% \begin{center}
% \begin{minipage}{\textheight}
% \caption{Tables which are too long to fit, should be written using the ``sidewaystable'' environment as shown here}\label{tab3}
% \begin{tabular*}{\textheight}{@{\extracolsep{\fill}}lcccccc@{\extracolsep{\fill}}}
% \toprule%
% & \multicolumn{3}{@{}c@{}}{Element 1\footnotemark[1]}& \multicolumn{3}{@{}c@{}}{Element\footnotemark[2]} \\\cmidrule{2-4}\cmidrule{5-7}%
% Projectile & Energy	& $\sigma_{calc}$ & $\sigma_{expt}$ & Energy & $\sigma_{calc}$ & $\sigma_{expt}$ \\
% \midrule
% Element 3 & 990 A & 1168 & $1547\pm12$ & 780 A & 1166 & $1239\pm100$ \\
% Element 4 & 500 A & 961  & $922\pm10$  & 900 A & 1268 & $1092\pm40$ \\
% Element 5 & 990 A & 1168 & $1547\pm12$ & 780 A & 1166 & $1239\pm100$ \\
% Element 6 & 500 A & 961  & $922\pm10$  & 900 A & 1268 & $1092\pm40$ \\
% \botrule
% \end{tabular*}
% \footnotetext{Note: This is an example of table footnote this is an example of table footnote this is an example of table footnote this is an example of~table footnote this is an example of table footnote.}
% \footnotetext[1]{This is an example of table footnote.}
% \end{minipage}
% \end{center}
% \end{sidewaystable}

\begin{table*}[ht]
\centering
\small
% \begin{tabular}{p{2cm} p{2.6cm} p{2.6cm} p{2.6cm} p{2.6cm}} \hline
% \begin{tabular}{r p{2.8cm} l l l} \hline
\resizebox{1.02\textwidth}{!}{
\begin{tabular}{r l l l l} \hline
                & RMSE              & $\downarrow$ 20\% Info      & 50:50                & $\uparrow$ 20\% Info \\ \hline
Base:Avg.       &   0.214 (0.206, 0.221)   & 0.500 (0.500, 0.500) & 0.500 (0.500, 0.500) & 0.500 (0.500, 0.500)\\
Base:BoW        &   0.193 (0.183, 0.203)   & 0.809 (0.772, 0.847) & 0.757 (0.728, 0.787) & 0.744 (0.714, 0.773)\\
Base:Length     &   0.180 (0.173, 0.187)   & 0.781 (0.762, 0.799) & 0.755 (0.736, 0.775) & 0.749 (0.725, 0.773)\\
Base:ELMo       &   0.179 (0.170, 0.188)   & 0.858 (0.828, 0.887) & 0.806 (0.784, 0.827) & 0.775 (0.748, 0.802)\\
Ours:ELMo+Att   &   0.166 (0.159, 0.173)   & 0.868 (0.838, 0.898) & 0.810 (0.787, 0.834) & 0.778 (0.749, 0.807)\\
Base:BERT       &   0.201 (0.192, 0.210)   & 0.806 (0.763, 0.849) & 0.743 (0.710, 0.775) & 0.707 (0.672, 0.743)\\
Ours:BERT+Att   &   \textbf{0.154 (0.146, 0.162)}   & \textbf{0.895 (0.875, 0.916)} & \textbf{0.842 (0.824, 0.860)} & \textbf{0.791 (0.773, 0.809)}\\\hline
\end{tabular}
}
\caption{
% BWS
Average RMSE and binary classification results (ROCAUC) with the single-sentence context dataset. 
For both ELMo and BERT-based models, adding the attention block (\texttt{+Att}) improved the performance for predicting continuous score range (\textit{RMSE}) and various classification scenarios (\texttt{ROCAUC}).
% ROCAUC scores compare different classification scenarios. All models performed significantly better performance than the random guessing ($>0.5 ROCAUC$). 
% Numbers in bold indicate the performance score is significantly different (italic: marginally different) from another model's result using the same pre-training embedding type. 
Numbers in parentheses are the 95\% confidence interval. 
Numbers in bold indicate the best performing model in each criteria. 
}
\label{tab:pred_dscovar}
\end{table*}

\subsection{Multi-Sentence Contexts}
\label{app:exp1_results_multi}
Table \ref{tab:pred_kapelner} shows more details of our models' performance on the multi-sentence context dataset. 

\begin{table*}[h]
\centering
\small
\resizebox{1.02\textwidth}{!}{
% \begin{tabular}{p{2.6cm} p{2.6cm} p{2.6cm} p{2.6cm} p{2.6cm}} \hline
\begin{tabular}{r l l l l} \hline
                    & RMSE                  & $\downarrow$ 20\% Info     & 50:50         & $\uparrow$ 20\% Info \\ \hline
Base:Avg.           & 0.173, (0.170, 0.176) & 0.500 (0.500, 0.500)  & 0.500 (0.500, 0.500)  & 0.500 (0.500, 0.500) \\
Base:BoW            & 0.201, (0.199, 0.204) & 0.643 (0.630, 0.656)  & 0.599 (0.588, 0.610)  & 0.585 (0.575, 0.595) \\
Base:Length         & 0.173, (0.170, 0.176) & 0.511 (0.505, 0.517)  & 0.507 (0.500, 0.514)  & 0.502 (0.495, 0.509) \\
Base:RF\_Lex        & 0.157, (0.154, 0.159) & 0.736 (0.729, 0.743)  & 0.698 (0.691, 0.705)  & 0.680 (0.669, 0.692) \\
Base:ELMo           & 0.152, (0.146, 0.159) & 0.768 (0.757, 0.779)  & 0.729 (0.721, 0.737)  & 0.705 (0.696, 0.715) \\
Ours:ELMo+Att       & 0.153, (0.149, 0.156) & 0.770 (0.760, 0.780)  & 0.727 (0.720, 0.734)  & 0.701 (0.689, 0.713) \\
Ours:ELMo+Att+Lex   & 0.152, (0.149, 0.155) & 0.789 (0.779, 0.799)  & 0.746 (0.739, 0.754)  & 0.725 (0.719, 0.731) \\
Base:BERT           & 0.139, (0.136, 0.142) & 0.807 (0.797, 0.817)  & 0.764 (0.757, 0.772)  & 0.751 (0.739, 0.763) \\
Ours:BERT+Att       & \textbf{0.138, (0.136, 0.140)} & 0.816 (0.806, 0.825)  & 0.777 (0.770, 0.785)  & 0.768 (0.757, 0.778) \\
Ours:BERT+Att+Lex   & 0.145, (0.142, 0.149) & \textbf{0.822 (0.814, 0.831)}  & \textbf{0.782 (0.775, 0.788)}  & \textbf{0.773 (0.765, 0.781)} \\ \hline
\end{tabular}
}
\caption{
Average RMSE and binary classification results (ROCAUC) with the multi-sentence context dataset~\citep{kapelner2018predicting}.
The BERT-based model with the attention block (\texttt{+Att}) performed significantly better than the baseline and ELMo-based models, in terms of RMSE and ROCAUC scores. 
Adding features from the original paper~\citep{kapelner2018predicting} (\texttt{+Lex}) also increased the prediction performance.
Numbers in parentheses are the 95\% confidence interval. 
Numbers in bold indicate the best performing model in each criteria. 
}
\label{tab:pred_kapelner}
\end{table*}

%%%%%%%%%%%%%%%%%%%%%%%%%%%%%%%%%%%%%%%%%%%%%%%%%%%%%%%%%%%%%%%%%%%%%%%%%%%%%%%%%%%%%%%%%%%%%%%%%%%%%%%%%%%%%%%%%%%

\subsection{EVALUtion dataset}
\label{app:exp2_evalution}
Table \ref{tab:evaluation_examples} shows the list of relations and templates that~\citep{santus2015evalution} used to create example sentences. 
Table \ref{tab:evaluation_scores_rdm} includes the randomized rank scores for each relation that used as a baseline for EVALution dataset~\citep{santus2015evalution}.
Table \ref{tab:evaluation_scores_single} shows the results from models trained with the single-sentence context data.
And Table \ref{tab:evaluation_scores_multi} shows the results from models trained with the multi-sentence context data.

\begin{table*}[t]
\centering
\small
% \begin{tabular}{r l l l}\hline
\begin{tabular}{r l l p{5cm}}\hline
Relation                & Pairs     & Relata    & Senence template \\ \hline
IsA (hypernym)          & 1880      & 1296      & X is a \textit{kind} of Y \\
Antonym                 & 1660      & 1144      & X can be used as the \textit{opposite} of Y\\
Synonym                 & 1086      & 1019      & X can be used with the \textit{same} meaning of Y\\
Meronym                 & 1003      & 978       & X is ...\\
- PartOf                & 654       & 599       & ... \textit{part} of Y\\
- MemberOf              & 32        & 52        & ... \textit{member} of Y\\
- MadeOf                & 317       & 327       & ... \textit{made} of Y\\
Entailment              & 82        & 132       & If X is \textit{true}, then also Y is \textit{true}\\
HasA (possession)       & 544       & 460       & X can \textit{have} or can \textit{contain} Y\\
HasProperty (attribute) & 1297      & 770       & Y is to \textit{specify} X\\ \hline
\end{tabular}
\caption{The list of relations, number of pairs, and sentence template examples from~\citep{santus2015evalution}.
In a sentence template, X means the target word, while Y is the pair, and italicized words are relational cues. 
}
\label{tab:evaluation_examples}
\end{table*}

\begin{table*}[t]
\centering
\small
\begin{tabular}{r l l}\hline
Relations       & \texttt{Rdm:pair}& \texttt{Rdm:rcue}\\ \hline
Antonym         & 0.507 (0.491, 0.523) & 0.494 (0.478, 0.510)\\
Entails         & 0.482 (0.414, 0.550) & 0.452 (0.380, 0.524)\\
HasA            & 0.485 (0.457, 0.513) & 0.507 (0.479, 0.535)\\
HasProperty     & 0.498 (0.479, 0.516) & 0.507 (0.489, 0.525)\\
IsA             & 0.508 (0.492, 0.523) & 0.502 (0.487, 0.517)\\
MadeOf          & 0.550 (0.512, 0.587) & 0.487 (0.446, 0.527)\\
MemberOf        & 0.425 (0.307, 0.543) & 0.494 (0.375, 0.612)\\
PartOf          & 0.501 (0.473, 0.529) & 0.476 (0.449, 0.502)\\
Synonym         & 0.493 (0.474, 0.512) & 0.503 (0.484, 0.522)\\
Overall Avg.    & 0.503 (0.495, 0.510) & 0.498 (0.491, 0.506)\\\hline
\end{tabular}
\caption{
Random baseline performance for EVALution dataset~\citep{santus2015evalution}. 
Numbers in parentheses are the 95\% confidence interval. 
}
\label{tab:evaluation_scores_rdm}
\end{table*}{}

\begin{table*}[t]
\centering
\small
\begin{tabular}{r l l}\hline
Relations       &   \texttt{ELMo+Att:pair} & \texttt{ELMo+Att:rcue}\\ \hline
Antonym         &  \textbf{0.727 (0.724, 0.730)}    & \textbf{0.603 (0.599, 0.606)} \\
Entails         &  0.364 (0.346, 0.382)             & 0.210 (0.183, 0.237)          \\
HasA            &  \textbf{0.509 (0.504, 0.515)}    & 0.257 (0.252, 0.261)          \\
HasProperty     &  0.405 (0.403, 0.408)             & \textbf{0.795 (0.794, 0.797)} \\
IsA             &  0.400 (0.397, 0.403)             & \textbf{0.795 (0.793, 0.796)} \\
MadeOf          &  0.504 (0.499, 0.510)             & 0.259 (0.253, 0.265)          \\
MemberOf        &  0.400 (0.400, 0.400)             & \textbf{0.800 (0.800, 0.800)} \\
PartOf          &  \textbf{0.504 (0.501, 0.507)}    & 0.250 (0.248, 0.252)          \\
Synonym         &  \textbf{0.771 (0.768, 0.774)}    & 0.477 (0.473, 0.481)          \\
Overall Avg.    &  \textbf{0.546 (0.542, 0.549)}    & \textbf{0.592 (0.587, 0.597)} \\\hline
\end{tabular}
\begin{tabular}{r l l}\hline
Relations       & \texttt{BERT+Att:pair}&\texttt{BERT+Att:rcue}\\ \hline
Antonym         & 0.353 (0.346, 0.359)              & \textbf{0.501 (0.494, 0.508)}\\ 
Entails         & 0.451 (0.415, 0.486)              & \textbf{0.514 (0.487, 0.540)}\\ 
HasA            & \textbf{0.527 (0.515, 0.540)}     & 0.430 (0.427, 0.433)\\ 
HasProperty     & 0.031 (0.027, 0.035)              & \textbf{0.592 (0.590, 0.595)}\\ 
IsA             & 0.024 (0.020, 0.027)              & 0.437 (0.432, 0.441)\\ 
MadeOf          & 0.007 (0.002, 0.012)              & \textbf{0.740 (0.734, 0.747)}\\ 
MemberOf        & 0.056 (0.014, 0.098)              & \textbf{0.569 (0.536, 0.601)}\\ 
PartOf          & 0.023 (0.017, 0.029)              & \textbf{0.711 (0.703, 0.719)}\\ 
Synonym         & \textbf{0.567 (0.557, 0.577)}     & \textbf{0.608 (0.600, 0.617)}\\ 
Overall Avg.    & 0.215 (0.209, 0.220)              & \textbf{0.540 (0.537, 0.543)}\\\hline
\end{tabular}
% }
\caption{
Comparing the average of normalized rank scores of attention weights from single-sentenced trained models across various relations in EVALution dataset~\citep{santus2015evalution}. 
Higher score means better.
A better model should pay more weights to the word that pairs with the target word (\texttt{pair}) or words that can indicate the type of relationship between the target and the pair word (\texttt{rcue}). 
The \texttt{ELMo+Att} model perform better than the \textit{BERT+Att} model in both \texttt{pair} and \texttt{rcue} scores. However, the performance is less accurate than multi-sentence dataset models in Table \ref{tab:evaluation_scores_multi}. 
Numbers in parentheses are the 95\% confidence interval. 
Numbers in bold indicate performance better than the random baseline ($=0.5$ in normalized rank score). 
}
\label{tab:evaluation_scores_single}
\end{table*}

\begin{table*}[t]
\centering
\small
\resizebox{1.02\textwidth}{!}{
% \begin{tabular}{p{2cm} p{2.6cm} p{2.6cm} p{2.6cm} p{2.6cm}} \hline
\begin{tabular}{r l l l l}\hline
Relations   &   \texttt{ELMo+Att:pair}  & \texttt{ELMo+Att:rcue}& \texttt{BERT+Att:pair}& \texttt{BERT+Att:rcue}\\ \hline
Antonym     & 	0.505 (0.504, 0.507)             & 0.253 (0.251, 0.254)         & 0.462 (0.451, 0.474)             & \textbf{0.599 (0.588, 0.610)}\\
Entails     & 	\textbf{0.954 (0.919, 0.989)}    & 0.394 (0.377, 0.411)         & \textbf{0.664 (0.616, 0.711)}    & \textbf{0.616 (0.585, 0.647)}\\
HasA        & 	0.310 (0.304, 0.317)             & \textbf{0.758 (0.755, 0.762)}& \textbf{0.538 (0.518, 0.559)}    & \textbf{0.828 (0.818, 0.837)}\\
HasProperty & 	\textbf{0.998 (0.996, 1.000)}    & 0.209 (0.205, 0.212)         & \textbf{0.562 (0.557, 0.566)}    & 0.234 (0.228, 0.240)\\
IsA         & 	\textbf{0.990 (0.987, 0.993)}    & 0.212 (0.209, 0.215)         & 0.230 (0.227, 0.234)             & \textbf{0.911 (0.904, 0.919)}\\
MadeOf      & 	\textbf{0.998 (0.995, 1.001)}    & \textbf{0.692 (0.681, 0.704)}& \textbf{0.974 (0.958, 0.989)}    & \textbf{0.741 (0.735, 0.746)}\\
MemberOf    & 	\textbf{0.994 (0.981, 1.006)}    & 0.200 (0.200, 0.200)         & \textbf{0.819 (0.747, 0.891)}    & 0.431 (0.386, 0.477)\\
PartOf      & 	\textbf{0.995 (0.992, 0.998)}    & \textbf{0.705 (0.697, 0.712)}& \textbf{0.975 (0.966, 0.983)}    & \textbf{0.740 (0.734, 0.746)}\\
Synonym     & 	\textbf{0.777 (0.775, 0.780)}    & 0.235 (0.231, 0.239)         & \textbf{0.771 (0.764, 0.778)}    & 0.160 (0.153, 0.167)\\
Overall Avg.& 	\textbf{0.808 (0.802, 0.814)}    & 0.328 (0.324, 0.333)         & \textbf{0.542 (0.535, 0.548)}    & \textbf{0.585 (0.577, 0.592)}\\ \hline
\end{tabular}
}
\caption{
Comparing the average of normalized rank scores of attention weights from multi-sentenced trained models across various relations in EVALution dataset~\citep{santus2015evalution}. 
Higher score means better.
Overall, the \texttt{ELMo+Att} performs better in capturing the \texttt{pair} words. The \texttt{BERT+Att} model is better in capturing the \texttt{rcue} words.
However, \texttt{BERT+Att} model captured more relationships than the random baseline in both \texttt{pair} and \texttt{rcue} words simultaneously.
Numbers in parentheses are the 95\% confidence interval. 
Numbers in bold indicate performance better than the random baseline ($=0.5$). 
}
\label{tab:evaluation_scores_multi}
\end{table*}

\begin{table*}[t]
\centering
\small
\resizebox{1.02\textwidth}{!}{
% \begin{tabular}{p{2cm} p{2.6cm} p{2.6cm} p{2.6cm} p{2.6cm}} \hline
\begin{tabular}{r l l l l}\hline
Relations   &   \texttt{BERT\_Early:pair} &   \texttt{BERT\_Early:rcue}          &   \texttt{BERT\_Mid:pair}          &    \texttt{BERT\_Mid:rcue} \\\hline
Antonym     & 	0.230, (0.220, 0.240)   &   \textbf{0.511, (0.508, 0.515)}   &   \textbf{0.748, (0.739, 0.757)}   &    0.392, (0.386, 0.397)\\
Entails     & 	0.362, (0.278, 0.447)   &   0.212, (0.182, 0.243)            &   0.274, (0.221, 0.327)            &    0.306, (0.271, 0.340)\\
HasA        & 	0.490, (0.469, 0.512)   &   0.485, (0.478, 0.491)            &   \textbf{0.667, (0.650, 0.683)}   &    0.368, (0.362, 0.374)\\
HasProperty & 	0.259, (0.244, 0.274)   &   \textbf{0.542, (0.536, 0.547)}   &   0.482, (0.468, 0.495)            &    0.086, (0.078, 0.095)\\
IsA         & 	0.034, (0.029, 0.038)   &   0.388, (0.385, 0.392)            &   \textbf{0.646, (0.638, 0.655)}   &    0.042, (0.037, 0.047)\\
MadeOf      & 	0.283, (0.268, 0.297)   &   0.036, (0.020, 0.052)            &   0.134, (0.113, 0.155)            &    0.216, (0.193, 0.238)\\
MemberOf    & 	0.181, (0.125, 0.237)   &   0.069, (0.034, 0.104)            &   0.444, (0.355, 0.533)            &    0.250, (0.202, 0.298)\\
PartOf      & 	0.068, (0.055, 0.081)   &   0.454, (0.443, 0.465)            &   0.426, (0.407, 0.445)            &    0.091, (0.077, 0.104)\\
Synonym     & 	0.194, (0.181, 0.207)   &   0.540, (0.537, 0.542)            &   \textbf{0.658, (0.642, 0.674)}   &    \textbf{0.539, (0.533, 0.546)}\\
Overall Avg.& 	0.189, (0.183, 0.194)   &   0.458, (0.454, 0.461)            &   \textbf{0.597, (0.591, 0.603)}   &    0.235, (0.230, 0.241)\\\hline
\end{tabular}
}
\small
\begin{tabular}{r l l}\hline
Relations   &   \texttt{BERT\_Later:pair} &   \texttt{BERT\_Later:rcue}\\\hline
Antonym     & 	\textbf{0.582, (0.570, 0.594)}   &   0.031, (0.027, 0.034) 	\\
Entails     & 	0.463, (0.404, 0.522)            &   0.296, (0.261, 0.331) 	\\
HasA        & 	\textbf{0.737, (0.720, 0.754)}   &   0.500, (0.493, 0.508) 	\\
HasProperty & 	0.468, (0.451, 0.485)            &   0.110, (0.101, 0.119) 	\\
IsA         & 	0.268, (0.260, 0.276)            &   0.215, (0.208, 0.221) 	\\
MadeOf      & 	0.180, (0.154, 0.206)            &   0.371, (0.349, 0.392) 	\\
MemberOf    & 	0.238, (0.151, 0.324)            &   0.331, (0.272, 0.391) 	\\
PartOf      & 	0.124, (0.112, 0.136)            &   0.177, (0.165, 0.189) 	\\
Synonym     & 	\textbf{0.555, (0.543, 0.568)}   &   0.068, (0.062, 0.073) 	\\
Overall Avg.& 	0.431, (0.425, 0.438)            &   0.161, (0.157, 0.166) 	\\\hline
\end{tabular}
\caption{
Comparing the average of normalized rank scores of attention weights from off-the-shelf BERT model's layers across various relations in EVALution dataset~\citep{santus2015evalution}. 
Higher score means better.
Only 6th layer of BERT performed better than the random rank score with capturing the \texttt{pair} words. 
Numbers in parentheses are the 95\% confidence interval. 
Numbers in bold indicate performance better than the random baseline ($=0.5$). 
}
\label{tab:evaluation_scores_bert}
\end{table*}

% \section{Contextual Informativeness Model}
% \label{sec:appendix_informativeness_hyperparams}
% For our deep learning based contextual informativeness model, we used root mean square error (RMSE) as a loss function. During training, we fine-tuned BERT's last encoding layer.
% After the validation process, we chose hyper-parameters as following: 
% \begin{itemize}
%     \item Batch size: 16
%     \item Iteration: 3
%     \item Learning rate: $3\mathrm{e}{-5}$
%     \item ReLU layer dimension: 256. 
%     \item Attention block layers dimension: 768 (same as BERT's output)
% \end{itemize}

% For the study, we used our replication of the random forest model from~\citep{kapelner2018predicting}. We followed the original paper's setting, including setting the number of estimators as 500, and bootstrapping sample size as 10000. We validate our comparison by replicating their original model results $R^2$=0.179 with very similar results $R^2$=0.177 in our replication.

% When lexical features from ~\citep{kapelner2018predicting} were used with contextual embedding features, we could observe a small performance increase (RMSE: 0.145, (0.142, 0.149),  $\downarrow$ 20\% Info.: 0.822 (0.814, 0.831), 50:50: 0.782 (0.775, 0.788), $\uparrow$ 20\% Info.: 0.773 (0.765, 0.781)).

%%%%%%%%%%%%%%%%%%%%%%%%%%%%%%%%%%%%%%%%%%%%%%%%%%%%%%%%%%%%%%%%%%%%%%%%%%%%%%%%%%%%%%%%%%%%%%%%%%%%%%%%%%%%%%%%%%%

\section{Results: Experiments 3 and 4}
\subsection{Contextual Informativeness Scores of Target Sentences}
\label{app:exp3__scores_and_sentences}
\begin{figure}[t]
\centering
% \hspace*{-0.61in}
\includegraphics[width=0.5\linewidth]{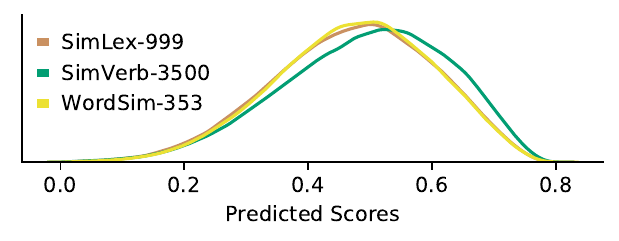}
\caption{
Distributions of predicted contextual informativeness scores for target sentences from ukWaC corpus. 
Predicted scores were centered around the center, with a sufficient number of relatively low- or high- informative sentences. %($\mu\approx0.5$, $\sigma\approx0.13$). 
}
\label{fig:curriculum_info_pred}
\end{figure}

Before Experiments 3 and 4, we briefly explored properties of target sentences. 
Figure \ref{fig:curriculum_info_pred} shows the distribution of the predicted scores from the contextual informativeness model. For all three semantic similarity tasks, contextual informativeness scores for target sentences were distributed around the score of 0.5, showing that our training data had a sufficient number of low or high informative sentences to test different curriculum building heuristics (standard deviation of $\sigma\approx0.13$). 
% , so we could say that our curriculum based on predicted informativeness scores were less biased in sentences' lengths. 

\begin{figure}[t!]
\centering
% \hspace*{-0.61in}
\includegraphics[width=0.5\linewidth]{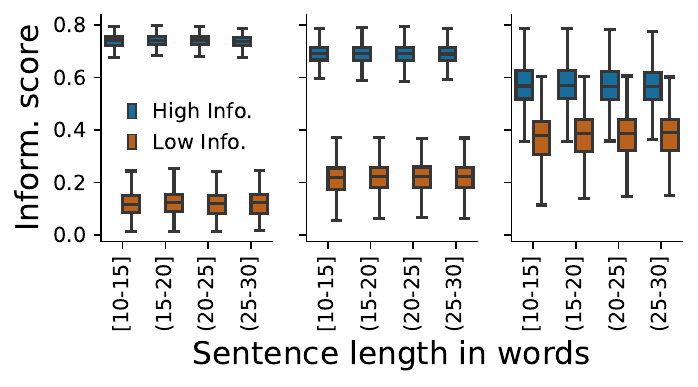}
\caption{
Relationship between sentence length and contextual informativeness predictions. 
Each plot represents when 4, 32, and 256 sentences are selected per target word. 
When the number of sentences selected per target words is small, the difference between high and low informative sentences' scores are more distinct.
Distributions of the predicted scores are not correlated to the length of sentences. 
}
\label{fig:curriculum_info_pred_simlex}
\end{figure}

% To ensure that a similar amount of lexical information was provided to the embedding models from different curricula, we ran the analysis in Figure \ref{fig:curriculum_info_pred_simlex}, which shows that sentence lengths and predicted informative scores of SimLex-999's target sentences in ukWaC corpus were not correlated.
We also wanted to ensure that a similar amount of lexical information was provided to the embedding models from different curricula. Figure \ref{fig:curriculum_info_pred_simlex} shows that sentence lengths and predicted informative scores of SimLex-999's target sentences in ukWaC corpus were not correlated.
Target sentences for SimVerb-3500 and WordSim-353 tasks also showed similar results.

\subsection{Word2Vec and FastText Models}
\label{sec:appendix_wordembeddings_hyperparams}
For consistent analysis results, we used the same hyper-parameters to train the background models for Word2Vec, FastText, and Nonce2Vec: 
\begin{itemize}
    \item Skip-gram algorithm 
    \item Embedding dimension: 400
    \item Window size: 5
    \item Negative sampling words: 5
    \item Minimum word counts: 50
    \item Alpha: 0.025
    \item Sampling rates: 0.001
\end{itemize}

When updating Word2Vec and FastText models, we changed the minimum word count to 0 for learning to accommodate learning with small-sized training sentences.

\subsection{Nonce2Vec Models}
\label{sec:appendix_nonce2vec_hyperparam}
We used Nonce2Vec models for the few-shot learning analysis. 
Compared to Word2Vec and FastText, Nonce2Vec had unique training process. 
We also tested more numbers of hyper-parameters for Nonce2Vec, since the model's results were much more sensitive to parameter settings. 

For updating Nonce2Vec models, we followed the settings from~\cite{herbelot2017high}, using 15 window words, 3 negative sampling words, 1 minimum word counts, and 10000 sampling rates. We also tested different learning rates \{0.5, 1, 2\}, and epochs \{1, 5\} for Nonce2Vec models. 

\begin{figure}[ht]
\centering
\includegraphics[width=0.8\linewidth]{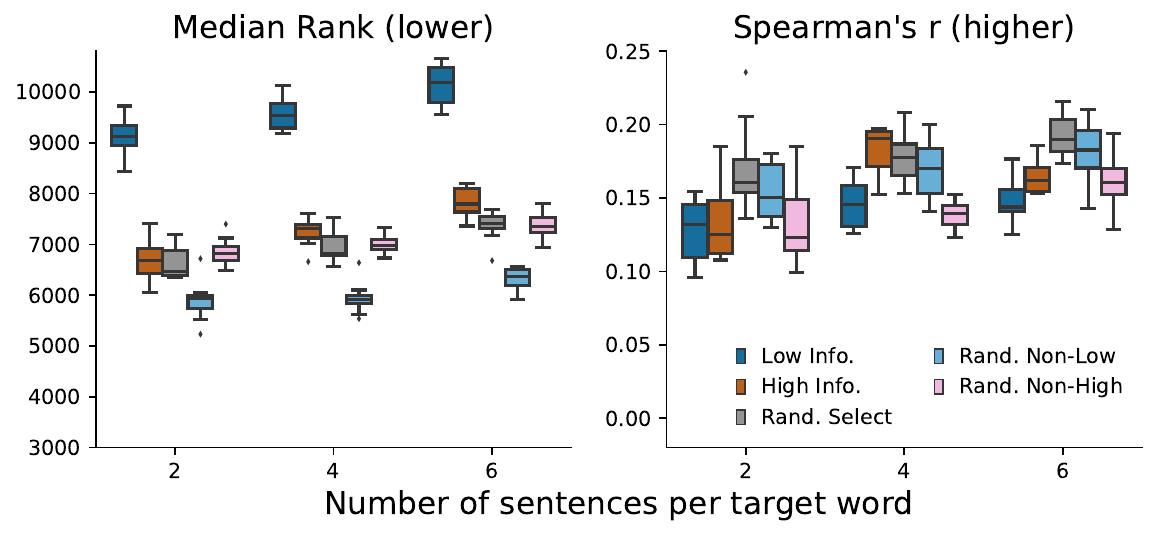}
\includegraphics[width=0.8\linewidth]{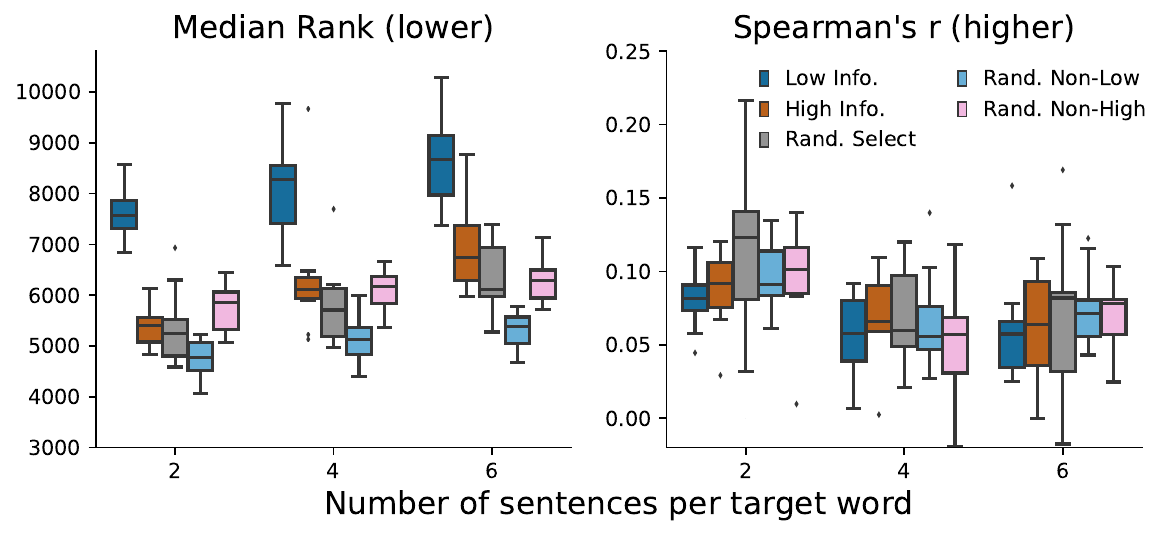}
\includegraphics[width=0.8\linewidth]{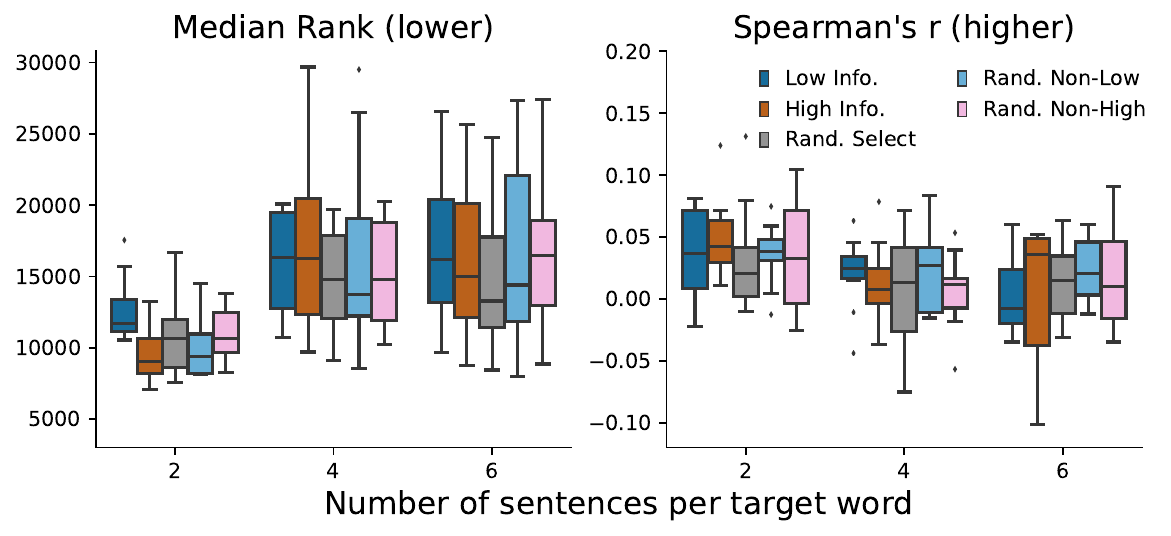}
\caption{
Few-shot learning results (Nonce2Vec) on SimLex-999 dataset, with different learning gains \{0.5, 1.0, 2.0\} and $epoch=1$ settings.
The highest learning rate setting showed the worst performance. 
}
\label{fig:curriculum_result_simlex_n2v_2_1}
\end{figure}

\begin{figure}[ht]
\centering
\includegraphics[width=0.8\linewidth]{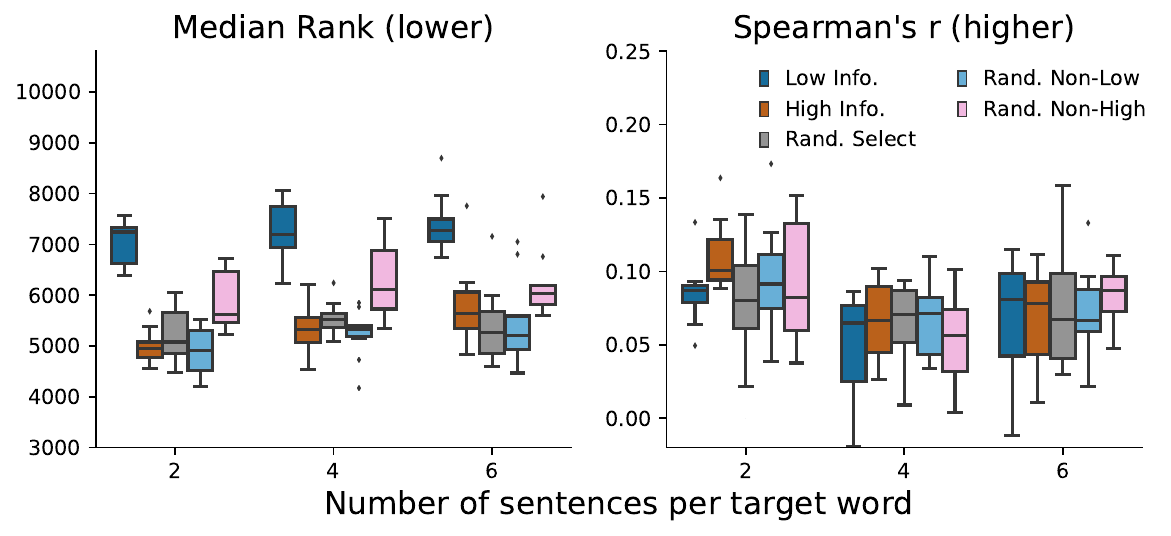}
\includegraphics[width=0.8\linewidth]{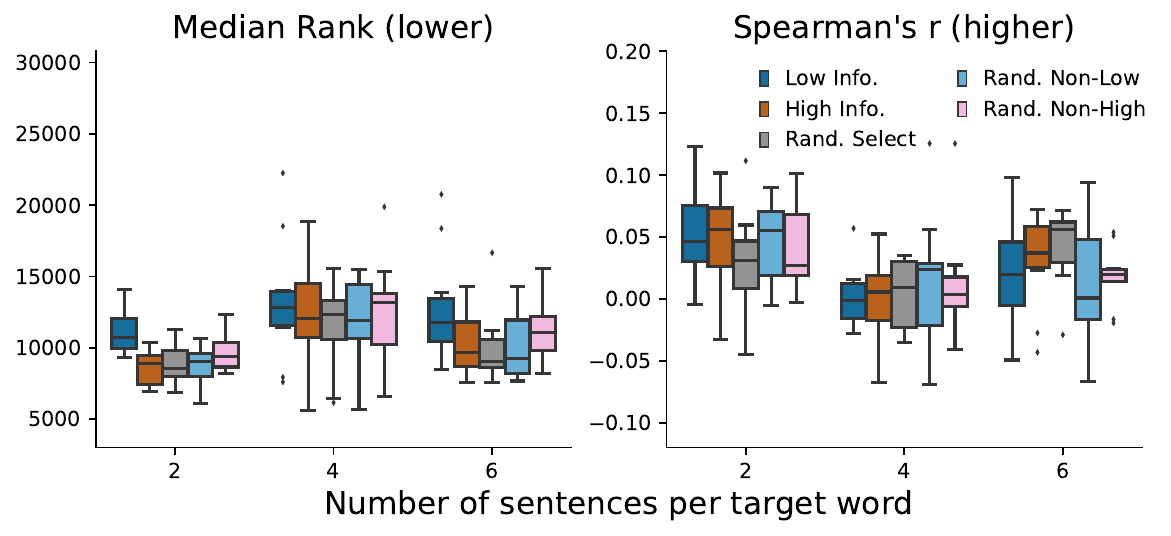}

\caption{
Few-shot learning results (Nonce2Vec) on SimLex-999 dataset, with different learning gains \{1.0, 2.0\} and $epoch=5$ settings.
Compared to $epoch=1$ settings, this higher epoch setting tended to show more stable and better results. 
}
\label{fig:curriculum_result_simlex_n2v_2_2}
\end{figure}

Figures \ref{fig:curriculum_result_simlex_n2v_2_1} and \ref{fig:curriculum_result_simlex_n2v_2_2} shows Nonce2Vec models' results in different hyper-parameter settings. 
Higher epoch setting (e.g., $epoch=5$) tended to show more stable results. 
The learning rate of 1.0 showed the best performance in median rank scores, but performed worse in Spearman's \textit{r} scores. 

\end{appendices}


\newpage
\bibliographystyle{plainnat}
\bibliography{main}% common bib file


\section*{Statements and Declarations}
This study was conducted as a part of the doctoral dissertation of Sungjin Nam while he was at the University of Michigan. 
% Research code and data will be available upon the paper’s acceptance.
% Some journals require declarations to be submitted in a standardised format. Please check the Instructions for Authors of the journal to which you are submitting to see if you need to complete this section. If yes, your manuscript must contain the following sections under the heading `Declarations':

% \begin{itemize}
% \item Conflict of interest/Competing interests (check journal-specific guidelines for which heading to use)
% \item Ethics approval 
% \item Consent to participate
% \item Consent for publication
% \item Availability of data and materials
% \item Authors' contributions
% \end{itemize}

% \noindent
% If any of the sections are not relevant to your manuscript, please include the heading and write `Not applicable' for that section. 

% %%===================================================%%
% %% For presentation purpose, we have included        %%
% %% \bigskip command. please ignore this.             %%
% %%===================================================%%
% \bigskip

% Springer journals and proceedings: \url{https://www.springer.com/gp/editorial-policies}

% Nature Portfolio journals: \url{https://www.nature.com/nature-research/editorial-policies}

% \textit{Scientific Reports}: \url{https://www.nature.com/srep/journal-policies/editorial-policies}

% BMC journals: \url{https://www.biomedcentral.com/getpublished/editorial-policies}
% \end{flushleft}

\end{document}